%% file: AiEDA.tex
\documentclass[journal, 10pt]{IEEEtran}

\input{ieda-config}

\usepackage{wasysym}
\usepackage[misc]{ifsym}

\title{AiEDA: An Open-Source AI-Aided Design Library for Design-to-Vector}

\author{Yihang~Qiu$^{1,2,*}$, 
        Zengrong Huang$^2$,
        Simin Tao$^2$, 
        Hongda Zhang$^1$,
        Weiguo Li$^2$,
        Xinhua Lai$^1$,
        Rui Wang$^3$, \\
        Weiqiang Wang$^1$, 
        Xingquan Li$^{2,~\textrm{\Letter}}$ \\
$^1$ University of Chinese Academy of Sciences, Beijing, China.\\
$^2$ Pengcheng Laboratory, Shenzhen, China.\\
$^3$ Shenzhen University, Shenzhen, China.\\
Email:  ~\textsuperscript{*}qiuyihang23@mails.ucas.ac.cn, ~\textsuperscript{~\textrm{\Letter}}lixq01@pcl.ac.cn
}

\begin{document}
\maketitle

\begin{abstract}
Recent research has demonstrated that artificial intelligence (AI) can assist electronic design automation (EDA) in improving both the quality and efficiency of chip design. But current AI for EDA (AI-EDA) infrastructures remain fragmented, lacking comprehensive solutions for the entire data pipeline from design execution to AI integration. Key challenges include fragmented flow engines that generate raw data, heterogeneous file formats for data exchange, non-standardized data extraction methods, and poorly organized data storage. This work introduces a unified open-source library for EDA (AiEDA) that addresses these issues. AiEDA integrates multiple design-to-vector data representation techniques that transform diverse chip design data into universal multi-level vector representations, establishing an AI-aided design (AAD) paradigm optimized for AI-EDA workflows. AiEDA provides complete physical design flows with programmatic data extraction and standardized Python interfaces bridging EDA datasets and AI frameworks. Leveraging the AiEDA library, we generate iDATA, a 600GB dataset of structured data derived from 50 real chip designs (28nm), and validate its effectiveness through seven representative AAD tasks spanning prediction, generation, optimization and analysis. The code is publicly available at \url{https://github.com/OSCC-Project/AiEDA}, {while the full iDATA dataset is being prepared for public release,} providing a foundation for future AI-EDA research.
\end{abstract}


\section{Introduction}
\label{sec:intro}

\IEEEPARstart{P}{hysical} implementation is an important part of electronic design automation (EDA), transforming gate-level netlists into manufacturable graphic design system (GDS-II) files. Recently, machine learning (ML) techniques have gained significant traction in EDA, demonstrating substantial potential for enhancing both efficiency and quality of physical design. These ML-based methods primarily target prediction\cite{duPowPrediCTCrossStagePower2024,wangMAUnetMultiscaleAttention2024a,parkPinAccessibilityRouting2024,ahnDTOCPDeepLearningDrivenTiming2024,chenPROS20PlugIn2023}, generation\cite{wuChatEDALargeLanguage2024,luGANPlaceAdvancingOpen2024a}, and optimization\cite{chenArbitrarysizeMultilayerOARSMT2024,zhangFastConstraintsTuning2024,yangMiracleMultiActionReinforcement2024} tasks across various design stages, including floorplanning\cite{yangMiracleMultiActionReinforcement2024}, placement\cite{luGANPlaceAdvancingOpen2024a}, clock tree synthesis\cite{ahnDTOCPDeepLearningDrivenTiming2024}, and routing\cite{chenArbitrarysizeMultilayerOARSMT2024}. Despite considerable progress in AI-aided design (AAD), the field still lacks specialized research infrastructure~\cite{kahngSolversEnginesTools2024a}. This gap has motivated the development of several AAD infrastructures, each addressing specific aspects of the research ecosystem.

\begin{figure}[!t]
\centering
\includegraphics[width=1.00\linewidth]{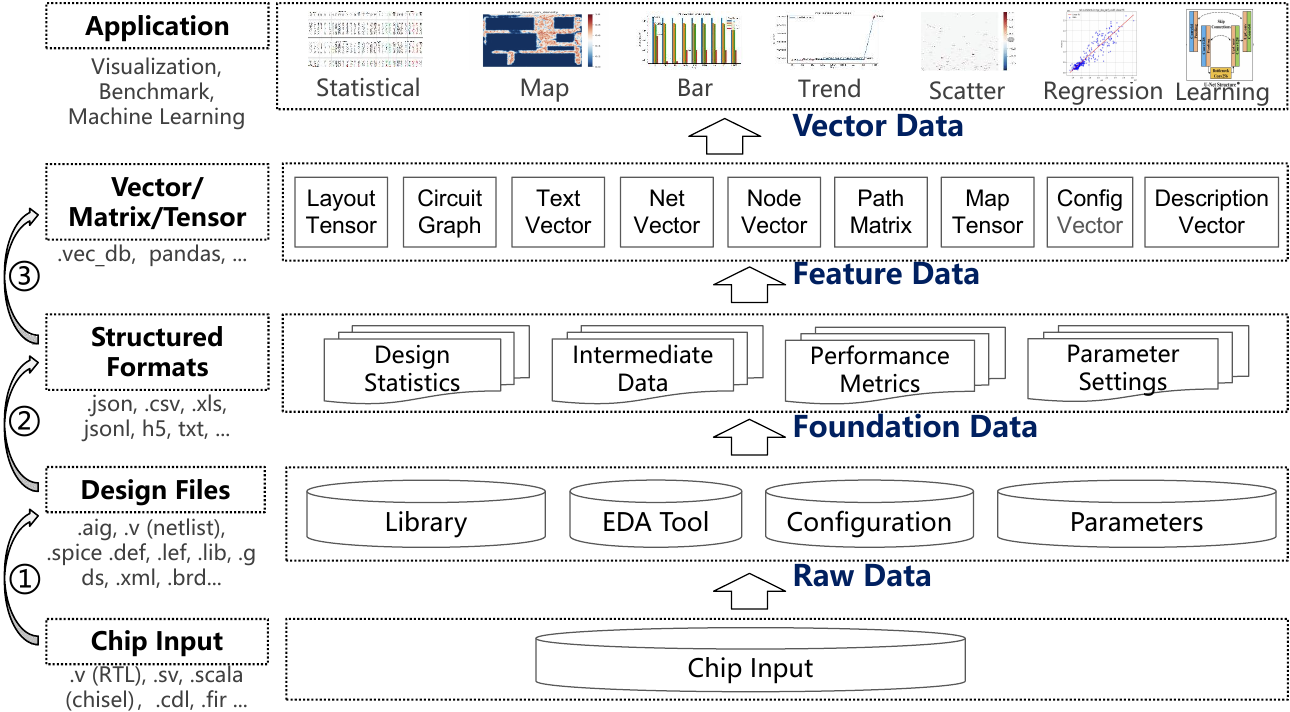}
\caption{{Data transformation pipeline for AI-aided design.}}
\label{fig:design-to-vector}
\end{figure}

\subsection{Available AI-EDA Infrastructures}
\label{sec:Available AI-EDA Infrastructures}

\idel{
To effectively advance AAD, establishing foundational infrastructures\idel{such as libraries and tools} is essential\idel{for enabling seamless conversion between chip design workflows and AAD's model training/inference processes}.
From a data perspective, the transformation pipeline involves {three}\idel{five} key stages: First, converting chip design data (\texttt{.verilog}) to design process data (\texttt{.netlist/.def}){; we define these original EDA tool outputs as \textbf{Raw Data}}. Second, extracting features and metrics into structured formats (\texttt{.json/.csv}). Third, transforming chip designs into vector/matrix/tensor representations—we refer to this stage as \textbf{design-to-vector} or \textbf{design vectorization}, where data preserving the original chip design information is called \textbf{source data}. Fourth, feeding structured and vectorized data into AI model training. Finally, integrating trained AI models into design flows to achieve AAD. }

{
To effectively advance AAD, establishing foundational infrastructures is essential. From a data perspective, the transformation pipeline involves three key stages. First, initial chip design data (\texttt{.verilog}) is converted into detailed design process data (\texttt{.netlist/.def/.gds}); we define these primary EDA tool outputs as \textbf{Raw Data}. Second, this Raw Data is parsed and structured into an AI-friendly, near-lossless representation that we term \textbf{Foundation Data} (e.g., organized into \texttt{.json/.csv} files). Inspired by the paradigm of Foundation Models in AI, this Foundation Data is engineered to be a clean, comprehensive, and reusable foundation for a diverse array of downstream AI tasks. Third, task-specific features (\textbf{Feature Data}) are extracted from the Foundation Data to generate the final vector, matrix, or tensor representations used directly by machine learning models. Collectively, we refer to these three stages as the \textbf{design-to-vector} pipeline.}
\cref{fig:design-to-vector} illustrates the complete data transformation pipeline for AAD. Several critical components are essential for this process: complete physical design flows with programmatic data generation capabilities, standardized methodologies for data representation and organization, and unified AI interfaces bridging EDA datasets and ML frameworks. These foundational elements collectively constitute AAD infrastructures, serving as the backbone for systematic and reproducible AI-driven EDA research.

\cref{tab:infra-comparison} compares existing open-source infrastructures for physical design across \idel{four}{three} dimensions: toolchain integration\idel{(multi vs. single)}, data extraction \idel{and type }support,\idel{design vectorization capabilities,} and AI interface availability. OpenLANE~\cite{shalanBuildingOpenLANE130nm2020a} and SiliconCompiler~\cite{siliconcompiler} integrate multiple EDA tools to automatically generate layouts from any register transfer level (RTL) design. Single-toolchain infrastructures such as OpenROAD~\cite{ajayiINVITEDOpenSourceDigital2019} and iEDA~\cite{liIEDAOpensourceInfrastructure2024,Li24iPD} provide Python interfaces that encapsulate their underlying C++ APIs, enabling rapid data extraction. METRICS2.1~\cite{jungMETRICS21FlowTuning2021a} supports flow execution using OpenROAD and standardizes design process metrics collection in JSON format. CircuitOps~\cite{liangInvitedPaperCircuitOps2023c} introduces an intermediate representation for netlists, constructing labeled property graphs and storing feature data in multiple CSV files to facilitate data preprocessing using Python libraries. Its integration into OpenROAD (CircuitOps+~\cite{chhabriaOpenROADCircuitOpsInfrastructure2024}) enhances ML interaction through Python API feedback loops. {Other related efforts structured design data for different goals, such as accelerating EDA tool~\cite{Fontana2017}, rather than for AI integration.} \idel{This comparison reveals that existing infrastructures target specific aspects, while AiEDA (this work) provides comprehensive coverage across all dimensions: support for multiple design flows (e.g., iEDA, OpenROAD, Innovus), comprehensive feature extraction interfaces, preservation of source data through vectorization methods, and unified AI interfaces bridging EDA datasets and ML frameworks.}{\cref{tab:infra-comparison} reveals that while pioneering infrastructures provide critical capabilities, they often target specific aspects. For instance, CircuitOps is expertly optimized for a graph-centric representation of netlists, while iEDA, is a self-contained physical design implementation, offers limited capabilities for AI integration. In contrast, AiEDA (this work) provides comprehensive coverage across all dimensions: support for multiple design flows (e.g., iEDA, OpenROAD, Innovus), enabling Foundation Data extraction, and unified AI interfaces bridging EDA datasets and ML frameworks.}

\subsection{Our Motivation and Contribution}
\label{sec:motivation}

\idel{Current fragmented AAD infrastructures primarily stem from several data challenges. To address these challenges, a key insight is converting chip design data into structured representations compatible with standard neural network model input/output pipelines. We refer to this data transformation process as design-to-vector shown in \cref{fig:design-to-vector}. We further define this data-oriented design paradigm as AI-aided design (AAD). Building upon this innovative data representation methodology, we have enhanced the capabilities of the AiEDA library, enabling it to effectively tackle the critical data challenges.} {The fragmentation of existing AAD infrastructures, highlighted in \cref{tab:infra-comparison}, underscores the need for a more systematic approach. Our key insight is that overcoming these limitations requires a robust \textbf{methodology} and \textbf{platform} for transforming design data into AI-ready formats. Building directly upon the design-to-vector paradigm, we developed AiEDA, a comprehensive library designed to serve as this unified foundation.} The contributions can be summarized as follows:

\begin{table}[t]
\centering
\scriptsize
\setlength{\tabcolsep}{4.4pt}
\caption{{Comparison of infrastructures for physical design.}}
\begin{tabular}{c|c|c|c}
\toprule
\diagbox{Work}{Function} & \makecell{Run Flow\\(Flow APIs)} & \makecell{Extract Data\\(Data APIs)} & \makecell{AI \\ Integration}  \\
\midrule
\makecell{OpenLANE \cite{shalanBuildingOpenLANE130nm2020a}} & Multi & $\times$   & $\times$  \\
\hline
\makecell{SiliconComliper \cite{siliconcompiler}} &Multi &$\times$ &  $\times$ \\
\hline
\makecell{OpenROAD \cite{ajayiINVITEDOpenSourceDigital2019}} &Single &\checkmark  & $\times$  \\
\hline
\makecell{iEDA \cite{liIEDAOpensourceInfrastructure2024,Li24iPD}} &Single &\checkmark  &   $\times$ \\
\hline
\makecell{METRICS2.1 \cite{jungMETRICS21FlowTuning2021a}} &Single &Design Process Metrics   &  $\times$  \\
\hline
\makecell{CircuitOps+ \cite{liangInvitedPaperCircuitOps2023c,chhabriaOpenROADCircuitOpsInfrastructure2024}} & Single & Graph-based Netlist Features & \checkmark   \\
\hline
\textbf{AiEDA}\textbf{(This work)} & Multi &  Foundation Data  &\checkmark  \\
\bottomrule
\end{tabular}
\label{tab:infra-comparison}
\end{table}


\begin{itemize}
    \item \idel{We propose the novel design-to-vector concept, enabling universal vectorized representation of heterogeneous chip design data. We define the AAD paradigm, shifting from traditional process-centric approaches to data-oriented workflows optimized for AI integration.}{We propose the \textbf{design-to-vector paradigm}, a systematic methodology for transforming heterogeneous chip design data into a variety of structured, AI-friendly formats (e.g., graphs, tensors, sequences). Based on this, we define the AAD paradigm, which shifts the focus from traditional process-centric approaches to data-oriented workflows optimized for AI integration.}
    
    \item We develop \textbf{AiEDA}, a comprehensive open-source AAD library featuring complete physical design workflows, programmatic data extraction, structured data management, and unified interfaces for AI applications.
    \idel{\item We release a ready-to-use dataset called \textbf{iDATA}, containing design-level, net-level, path-level, and patch-level information from 50 real chip designs, totaling 600GB.}
    \item {Leveraging our AiEDA library, we present \textbf{iDATA}, a ready-to-use 600GB dataset featuring multi-level structured data derived from 50 real 28nm chip designs.}
    
    \item We implement five representative tasks to demonstrate the effectiveness of our library and {structured} dataset, covering prediction, generation, and optimization\idel{across design-level, net-level, path-level, and patch-level representations}{, each targeting a distinct level of its representation hierarchy}.
\end{itemize}

This paper is structured as follows:
\cref{sec:background} details data challenges and examples of the design-to-vector concept. 
\cref{sec:flow} presents the AiEDA library architecture.
\cref{sec:dataset} describes the iDATA dataset.
\cref{sec:tasks} presents five downstream tasks with experimental results.
\cref{sec:conclusion} draws conclusions.

\section{Data Challenges and Design-to-Vector}
\label{sec:background}


\subsection{Data Challenges for AI-aided Design}
\label{sec:data challages for}
To enable AI-aided design (AAD), high-quality training data from real chip designs is crucial. However, current AAD infrastructure remains fragmented, with no complete solution existing for the entire data pipeline. Key challenges include: (1) fragmented flow engines that generate raw data, (2) inconsistent file formats for data exchange, (3) non-standardized data extraction methods, and (4) poorly organized data storage.



\subsubsection{Challenge 1: Fragmented Flow Engines}  
AI training data generation from RTL to GDS involves multiple stages and heterogeneous EDA tools, both commercial and open-source~\cite{parkPinAccessibilityRouting2024,chenPROS20PlugIn2023,luGANPlaceAdvancingOpen2024a}. Current tools often have inconsistent objectives and output diverse formats and metrics, leading to fragmented and low-quality data. This fragmentation lowers data quality, limits interoperability, and hinders model deployment in real design flows. A unified workflow platform is needed to standardize configurations, ensure consistency, and enable seamless AI integration for \idel{reliable} inference and optimization.

\subsubsection{Challenge 2: Heterogeneous File Formats}  
\cref{tab:input_file_formats} summarizes common input formats for physical design tools, illustrating file diversity. Each format possesses distinct syntax, semantics, and structures. These formats are incompatible with direct AI training, requiring extraction and transformation into vectorized representations\idel{such as CSV or NumPy arrays}. Without a unified conversion library, preprocessing becomes a bottleneck, delaying dataset creation and reducing reliability. Developing a standardized file-to-vector tool is essential to preserve design intent and relationships while enabling rapid, consistent data preparation for AI models, supporting scalable and effective automated design workflows.


\begin{table}[t]
\centering
\caption{File formats used by physical design tools.}
\label{tab:input_file_formats}
\begin{tabular}{|p{2.cm}|p{6cm}|}
\hline
\textbf{Function} & \textbf{Input} \\
\hline
Physical Implementation & 
 \texttt{.v} (Verilog), \texttt{.lef} (Library Exchange Format), \texttt{.def} (Design Exchange Format), \texttt{.lib} (timing library), \texttt{.sdc} (Synopsys Design Constraints) \\
\hline
Layout Design & 
\texttt{.cdl} (Circuit Description Language), \texttt{.spice} (simulation netlist), \texttt{.oa} (OpenAccess database), \texttt{.gds} (GDSII stream format) \\
\hline
Power Analysis & 
\texttt{.lef}, \texttt{.def}, \texttt{.lib}, \texttt{.upf/.cpf} (power intent), \texttt{.vcd} (switching activity), \texttt{.saif} (switching activity), \texttt{.captable} (parasitic tables) \\
\hline
Thermal Analysis & 
\texttt{.xml} (thermal parameters), \texttt{.csv} (thermal boundary conditions), \texttt{.flp} (floorplan), \texttt{.ptrace} (power trace) \\
\hline
Timing Analysis & 
\texttt{.sdc}, \texttt{.lib}, \texttt{.spef} (parasitic extraction), \texttt{.mmmc} (multi-mode multi-corner) \\
\hline
Design Rule Check (DRC) & 
\texttt{.gds}, \texttt{.lef}, \texttt{.rule/.drc} (DRC rules), \texttt{.layermap} (layer mapping) \\
\hline
Signal Integrity & 
\texttt{.spef}, \texttt{.lib}, \texttt{.sdc}, \texttt{.def}\\
\hline
\end{tabular}
\end{table}

\begin{figure*}[t]
    \centering
    \includegraphics[width=0.95\linewidth]{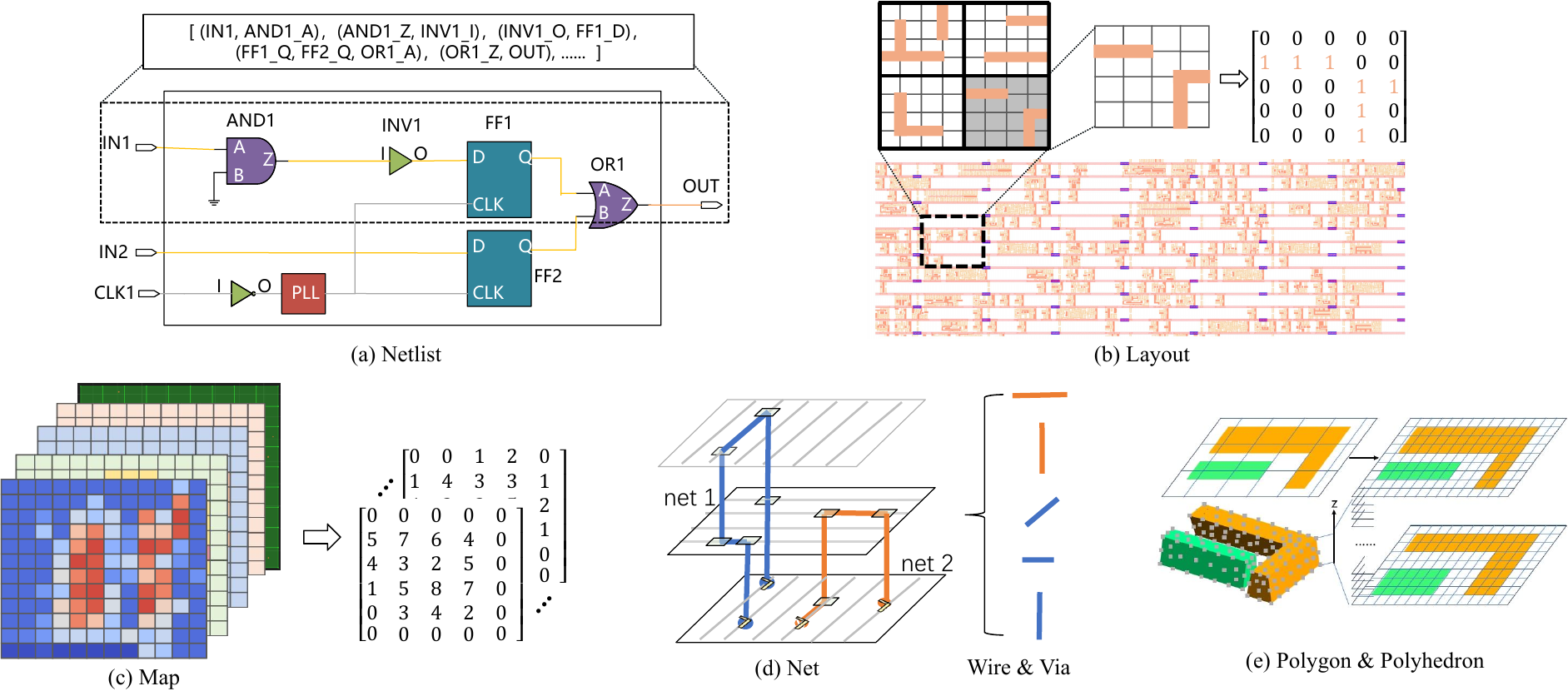}
    \caption{Examples of design-to-vector conversion for different data types.}
    \label{fig:component2vec}
\end{figure*}

\subsubsection{Challenge 3: Non-standardized Data Extraction}

AI-EDA research suffers from inconsistent data extraction methods, making results irreproducible. Different studies use diverse datasets, features, and parsing approaches, requiring custom tool interfaces and manual feature engineering. This inconsistency increases engineering effort, reduces feature reliability, and prevents fair benchmarking. The lack of standardization raises entry barriers and introduces variability in metrics\idel{, and hinders reproducibility}. A unified data extraction framework with standardized interfaces and verification mechanisms is essential to convert raw EDA data into consistent, AI-ready formats.


\subsubsection{Challenge 4: Chaotic Data Organization}

{Feature data from EDA tools is often stored in fragmented formats like CSV or NumPy, }\idel{EDA data are often fragmented in CSV or NumPy formats, }losing critical information and limiting reuse. Researchers repeatedly rebuild datasets, reducing efficiency and reproducibility. A unified framework is needed to preserve data integrity, support multi-tool interoperability, enable Python integration, and standardize access. This ensures consistent, AI-ready datasets, facilitates reproducible research, and accelerates AI-EDA development, overcoming inefficiencies caused by disorganized data storage and fragmented workflows.


\subsection{Solutions of Design-to-Vector}
\label{sec:solutions}
AI-EDA tasks can be simply classified into feature-based perception tasks and source data-based pre-training tasks. Most existing works focus on feature-based perception tasks. Beyond extracting features and metrics, we convert chip design data into structured representations (vectors, matrices, or tensors) compatible with neural network input/output pipelines. The following examples illustrate this conversion process. To address the data challenges, we introduce the \textbf{design-to-vector paradigm}. While converting design data for machine learning is not new, our paradigm formalizes and unifies prior task-specific efforts~\cite{renMachineLearningApplications2022} into a general framework. Crucially, ``vector" is used broadly to mean any structured, AI-ready data representation, explicitly including graphs for GNNs and images/tensors for CNNs. Our framework standardizes the conversion of raw, heterogeneous EDA data into these diverse, analysis-ready formats. This systematically lowers the engineering barrier for researchers, enabling them to readily generate graphs from netlists, tensors from layouts, and sequences from timing paths. The following examples illustrate how this paradigm is applied to different types of design data.

\begin{figure*}[t]
    \centering
    \includegraphics[width=0.98\linewidth]{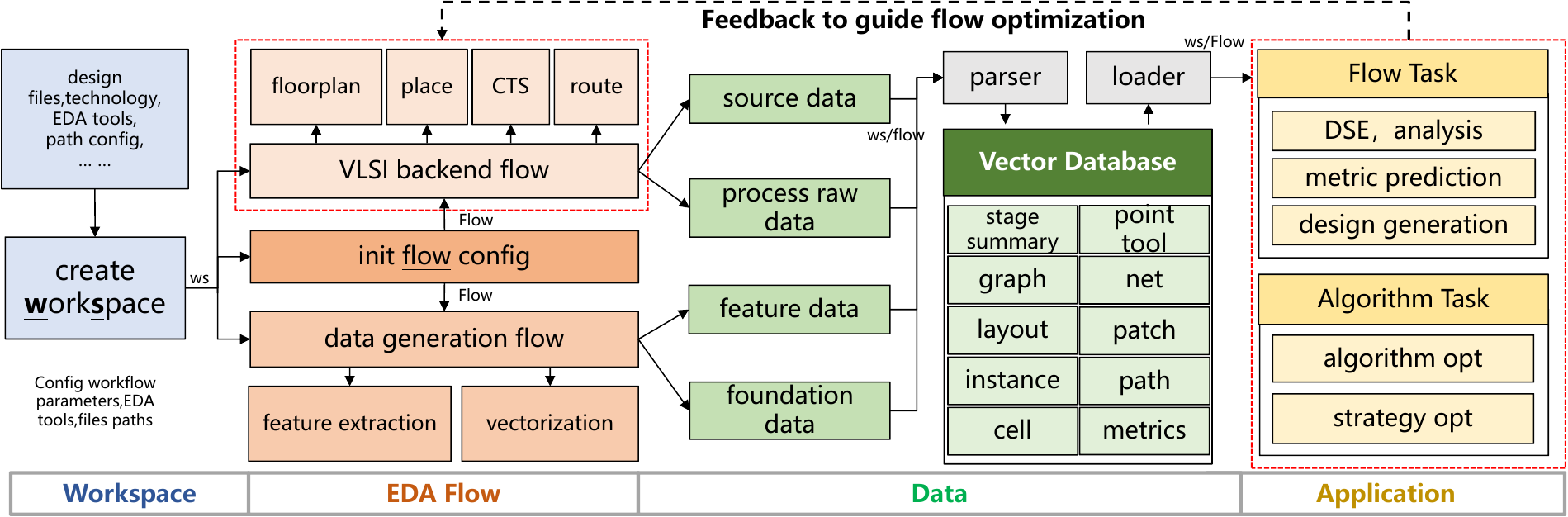}
    \caption{{The AI-Aided Design library for design-to-vector.}}
    \label{fig:library}
\end{figure*}

\subsubsection{Netlist-to-Vector}

A gate-level netlist consists of functional gates (e.g., AND, OR), sequential elements (e.g., flip-flops, latches), and their interconnecting nets. This structure can be abstracted as a hyper-graph \( G = (V, E) \), where \( V \) represents vertices (gates and primary I/Os) and \( E \) denotes hyper-edges (nets).  
For efficient storage, sparse representations such as the Sparse Adjacency Matrix and Incidence List are employed. 
\cref{fig:component2vec}(a) illustrates converting netlists (graph or path) to vectors, enabling AI tasks like gate classification, timing/power prediction, and performance optimization.

\subsubsection{Layout-to-Vector}  
A physical layout, represented by patterned layers (e.g., metal, via), can be treated as multi-channel image data. Discretization is achieved via Binary Pixelization or Multi-channel Encoding.  
\cref{fig:component2vec}(b) shows how geometric patterns are converted into vector representations. Layout and patch-level vectors support AI-EDA tasks such as congestion evaluation, DRC violation prediction and hotspot detection.

\subsubsection{Map-to-Vector}  
Physical design produces evaluation metrics such as timing, power, DRC, congestion, and IR drop maps, reflecting the spatial layout distribution. These metrics form grid-based structures that can be discretized into matrices.  
\cref{fig:component2vec}(c) shows this transformation into structured tensors. Map-level discretization enables multi-channel convolution for joint spatial analysis, supporting predictive modeling for hotspot mitigation, design closure acceleration, and Pareto-optimal optimization across electrical, thermal, and mechanical domains.

\subsubsection{Net-to-Vector}  
During routing, interconnects from the netlist are instantiated across metal and via layers following design rules. Each net can be decomposed into geometric primitives—wires and inter-layer vias using Steiner or critical points as anchors. Metal wires are represented as vectors \(\vec{w} = (x_s, y_s, x_e, y_e, l)\) and inter-layer vias are encoded as \(via = (x_c, y_c, l_b, l_t)\).  
\cref{fig:component2vec}(d) shows nets converted into discrete wires and vias. This representation supports AI tasks such as wirelength estimation, RC prediction, and rule-compliant routing generation.

\subsubsection{Shape-to-Vector}  
The final chip implementation consists of 2D and 3D shapes representing device structures. These are discretized into machine-readable forms while preserving topological features.  
2D shapes are rasterized into grids encoding material properties, with adjustable resolution balancing accuracy and storage.  
3D structures use techniques such as 2.5D layered grids, finite-element meshing, and point cloud sampling.  
\cref{fig:component2vec}(e) illustrates planar and 2.5D discretization. This representation supports AI-aided tasks like parasitic extraction, thermal-structural co-simulation, electromigration analysis, and manufacturability-aware optimization.

In AiEDA, we implement some design-to-vector approaches to convert complex design data into structured formats, efficiently address the key data challenges. To demonstrate this in practice, a complete, vectorized instance of the \texttt{gcd} design is available for inspection in our open-source repository\footnote{\url{https://github.com/OSCC-Project/AiEDA/tree/master/example/sky130\_gcd/output/iEDA/vectors}}.


\section{AiEDA: AI-Aided Design library}
\label{sec:flow}


Fig.~\ref{fig:library} illustrates the AAD library architecture for design-to-vector. The library comprises four essential components: 1) Flow Engines: supporting complete physical design workflows with flexible engine switching capabilities; 2) Data Generation: enabling programmatic control for flow execution and rich data extraction, facilitating efficient data acquisition; 3) Data Management: organizing workspaces for complex tool interactions and enabling batch vectorization through structured data organization; 4) Downstream Applications: providing unified interfaces for feature engineering and AI-based analysis, streamlining model training, validation, and evaluation. The following subsections detail these components.

\subsection{Flow Engines}
\label{subsec:flow_engines}
Flow engines form the foundational component for implementing and evaluating physical design processes. We categorize them into \textbf{tool engines} (implementing specific steps like placement and routing) and \textbf{evaluation engines} (assessing design features and performance metrics  through processes like static timing analysis). Available engines include open-source options (e.g., OpenROAD, iEDA) and commercial alternatives (e.g., Innovus, PrimeTime). Additionally, specialized tools like DREAMPlace\cite{linDREAMPlaceDeepLearning2021b} for placement and CUGR\cite{liuCUGRDetailedRoutabilityDriven3D2020} for routing often achieve superior results for specific steps.

Our library integrates common open-source engines, commercial tools, and specialized point tools. This integration relies on standardized data exchange formats (e.g., \texttt{.def}) to enable flexible tool switching—for instance, using DREAMPlace for placement while reverting to Innovus for routing. This eliminates integration complexities and lowers barriers for AI-EDA research, significantly expanding possibilities for comparative analysis. As most engines are implemented in C++, Python interface wrappers are required to ensure seamless integration within AAD environments. AiEDA integrates these diverse EDA tools as third-party libraries, managing and invoking them through a standardized and unified interface.

To facilitate intelligent and interpretable workflows in AI-assisted electronic design automation, the AiEDA library integrates two essential modules: the \texttt{aieda.report} and \texttt{aieda.gui} packages. These components together support both process-level insight and human-in-the-loop interaction, enabling designers to monitor, analyze, and present complex chip design tasks efficiently.

\begin{figure}
\centering
\includegraphics[width=1\linewidth]{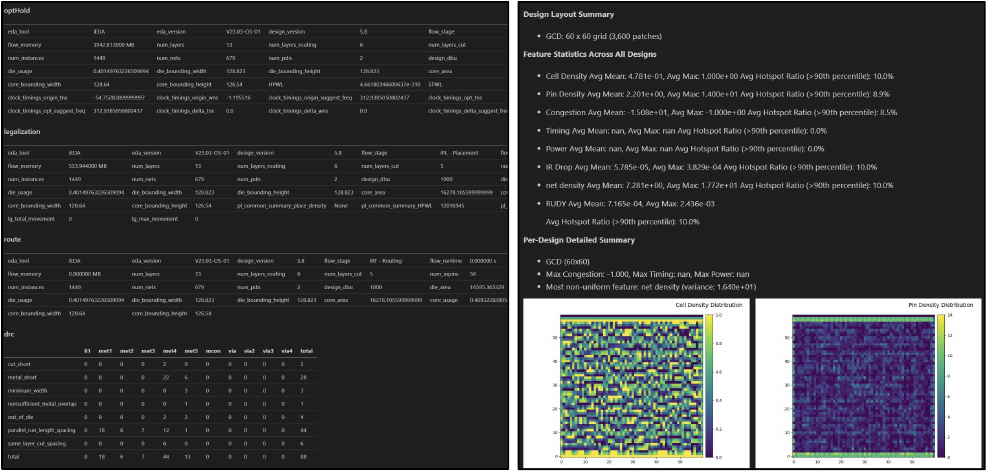}
\caption{AiEDA.reporter.}
\label{fig:report}
\end{figure}

The process and result reporter (\texttt{aieda.report}) serves as the analytical backbone of AiEDA, as shown in \cref{fig:report}. It automatically aggregates data from design stages, performs statistical analyses, and generates comprehensive, data-driven reports. Through the \texttt{ReportGenerator} class, users can produce customized documents that include performance metrics, comparison tables, and trend visualizations. The module also provides a \texttt{ReportModule} base class for extending reporting capabilities to specific EDA scenarios such as placement quality evaluation, timing closure verification, or power estimation. To enhance interpretability, the \texttt{VisualSummary} utility embeds concise graphical summaries directly within the report, allowing quick assessment of optimization results and anomalies. Reports can be exported to multiple formats (e.g., \texttt{PDF}, \texttt{HTML}), ensuring compatibility with existing documentation systems and collaborative workflows.

\begin{figure}
\centering
\includegraphics[width=1\linewidth]{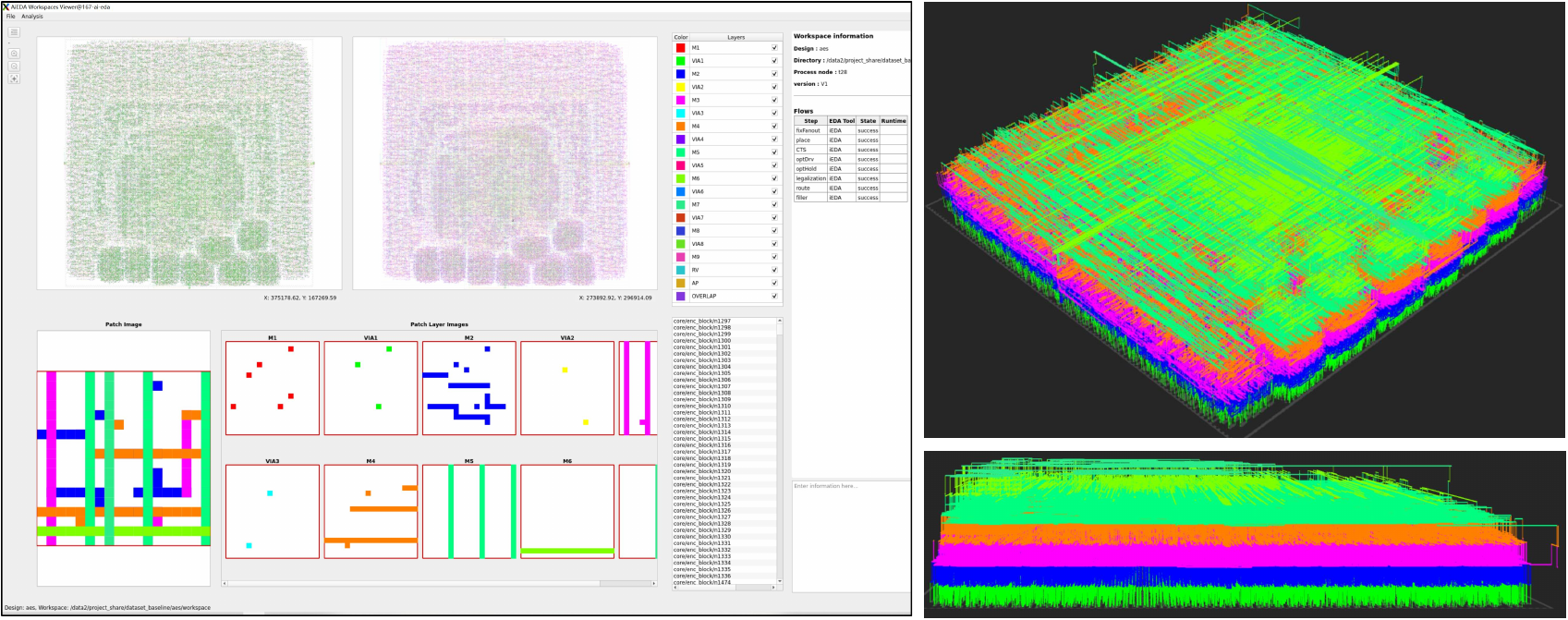}
\caption{AiEDA.GUI.}
\label{fig:gui}
\end{figure}

Complementing this, the process data and local visualizer (\texttt{aieda.gui}) provides an interactive graphical interface for real-time exploration of chip design data, as shown in \cref{fig:gui}. The module combines high-level design overviews with fine-grained local visualization capabilities. The \texttt{LayoutViewer} enables users to interactively inspect layout elements—cells, nets, and routing paths—with full zoom and pan support. For detailed physical inspection, the \texttt{ChipViewer} offers multi-layer visualization, revealing the geometric and material characteristics of chip components. The \texttt{PatchesViewer} introduces a grid-based visualization of local design regions, supporting tasks such as hotspot analysis and density distribution mapping. Additionally, the \texttt{WorkspaceManagerUI} provides interfaces for managing design workspaces, configurations, and versioned experiments, enhancing the reproducibility of AI-based EDA research.

\subsection{Data Generation}
\label{subsec:data_generation}

We encapsulate flow engines with Python interfaces for systematic data generation. The library provides two complementary APIs: \textbf{flow APIs} for design implementation and \textbf{data} for data extraction. These interfaces support both fine-grained operations (e.g., legalization, wirelength calculation) and coarse-grained workflows (e.g., full placement, comprehensive evaluation). 

Listing~\ref{lst:python_interface} demonstrates the unified data generation approach. Flow APIs enable flexible tool selection through input parameters—for instance, placement can utilize either iEDA's iPL or DREAMPlace. {Data} APIs specify evaluation tools and target design stages to extract stage-appropriate data. For example, for wirelength metrics, the system computes half-perimeter wirelength (HPWL) and rectilinear Steiner minimum tree (RSMT) during placement, while routed wirelength (RWL) is calculated after routing. For the same metric, multiple tools may be employed; for instance, DRC evaluation can employ either iEDA's iDRC or Innovus's verification interface.

\begin{figure}[htb]
\small
\begin{lstlisting}[style=pythonstyle, caption={{Unified Python API usage for data generation.}}, label=lst:python_interface]
from aieda.workspace import workspace_create
from aieda.flows import RunFlow
from aieda.data import RunFeature

def generate_data(ws_dir, tool, step):
    ws = workspace_create(ws_dir, tool)
    match step:
        case "floorplan":
            flow = RunFlow.runFP(ws, tool)
            feat = RunFeature.fp(ws, tool)
        case "place":
            flow = RunFlow.runPL(ws, tool)
            feat = RunFeature.pl(ws, tool)
        case "cts":
            flow = RunFlow.runCTS(ws, tool)
            feat = RunFeature.cts(ws, tool)
        case "routing":
            flow = RunFlow.runRT(ws, tool)
            feat = RunFeature.rt(ws, tool)
        case _:
            flow = RunFlow.run(ws, tool)
            feat = RunFeature.flow(ws, tool)
    return flow, feat
\end{lstlisting}
\end{figure}

Implementation complexity varies significantly across tool types. Open-source engines require only C++/Python linking via \texttt{pybind11}, while commercial tools demand TCL script generation, execution, and report parsing---substantially increasing implementation overhead.

Regarding data output, flow APIs generate standardized design files (e.g., \texttt{.def/.v}). {For comprehensive coverage, data APIs extract a wide range of information, spanning basic design statistics (e.g., cell counts, layout area), scalar metrics (e.g., wirelength, timing, power), and spatial data (e.g., congestion map, DRC hotspot).}{data} APIs extract engineered features, metrics, and source data for comprehensive data coverage. The extracted data span basic design statistics (e.g., cell counts, layout area), scalar metrics (e.g., wirelength, timing, power), and spatial data (e.g., congestion map, DRC hotspot). Interface invocation relies on pre-configured environment parameters for input/output paths, making comprehensive data management essential for multi-tool workflows.


\begin{figure}[!t]
\centering
\subfigure[]{
\includegraphics[width=0.99\linewidth]{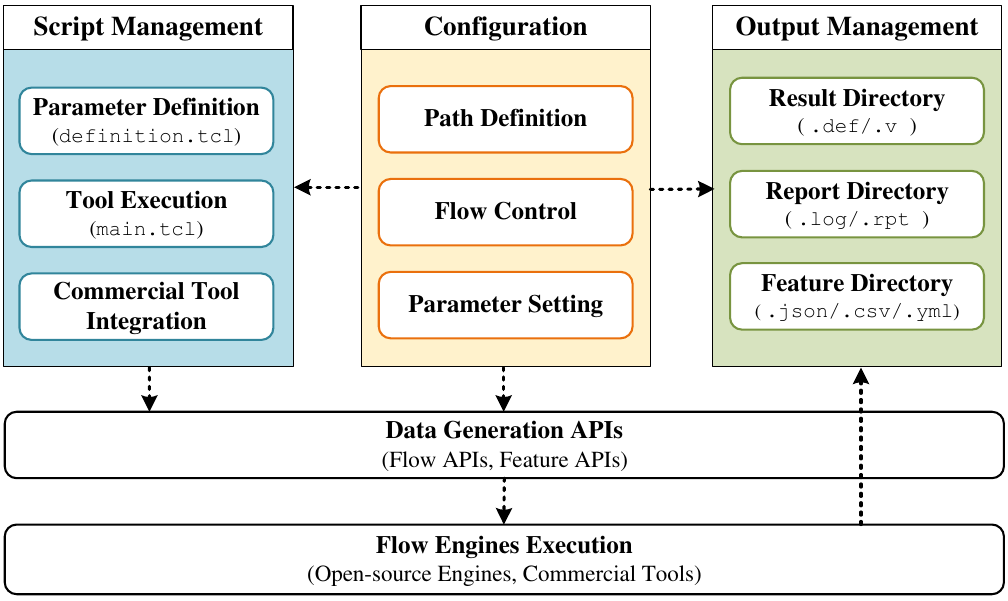}
}
\hfill
\subfigure[]{
\includegraphics[width=0.99\linewidth]{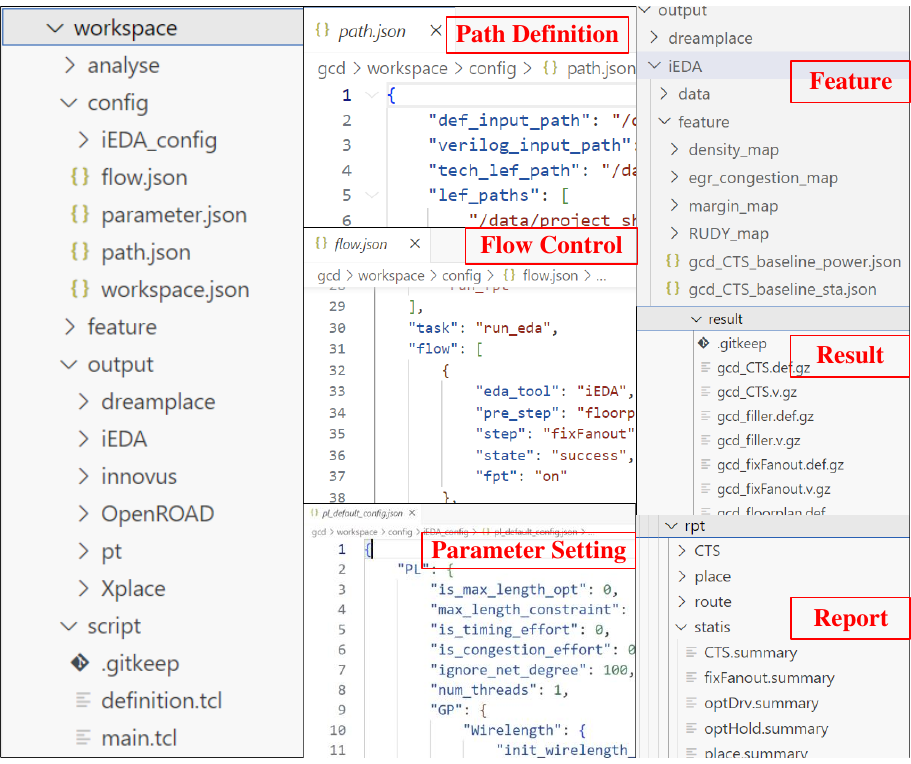}
}
\caption{Workspace architecture and file organization. (a) Core components and data flow. (b) File distribution.}
\label{fig:workspace}
\end{figure}

\subsection{Data Management} 
\label{subsec:data_management}

We propose two core concepts for comprehensive data management: \textbf{workspace} and \textbf{vectorization}. A workspace serves as a pre-configured environment that centralizes parameter settings, data access, and storage operations for each chip design. Vectorization transforms chip design data into structured representations compatible with standard neural network model input/output piplines. These enable standardized data management and high-throughput data generation through flexible Python interfaces.

\subsubsection{Workspace}

Fig.\ref{fig:workspace}(a) illustrates the three core workspace components, where dashed arrows indicate data flow relationships. Fig.\ref{fig:workspace}(b) demonstrates the hierarchical file organization within a workspace. The three core components are detailed as follows:

\paragraph{Configuration} serves as the workspace core, managing path definitions for all input/output files (e.g., \texttt{.lef/.def/.lib}) and providing centralized control over flow engines, including engine selection and execution parameters. Tool-specific configurations (e.g., ``\textit{target\_density}" for iEDA's iPL) are integrated through JSON files, ensuring interactive flexibility and workflow reproducibility.

\paragraph{Output Management} automatically routes generated data to corresponding output paths based on the selected engine. Data is systematically categorized into: (1) design files \texttt{.def/.v} stored in the result directory; (2) runtime files \texttt{.log/.rpt} archived in the report directory; and (3) extracted data \texttt{.json/.csv} organized in the feature directory. Users specify only the workspace root path while the system automatically handles file routing and organization.

\paragraph{Script Management} maintains execution scripts for commercial tool integration. When evaluating DRC with Innovus, the system generates TCL scripts comprising parameter definitions (\texttt{definition.tcl}) and tool execution commands (\texttt{main.tcl}). Parameter scripts configure input/output paths while execution scripts source these parameters and invoke tool-specific commands such as \texttt{verify\_drc}.

\begin{figure}
\centering
\includegraphics[width=1\linewidth]{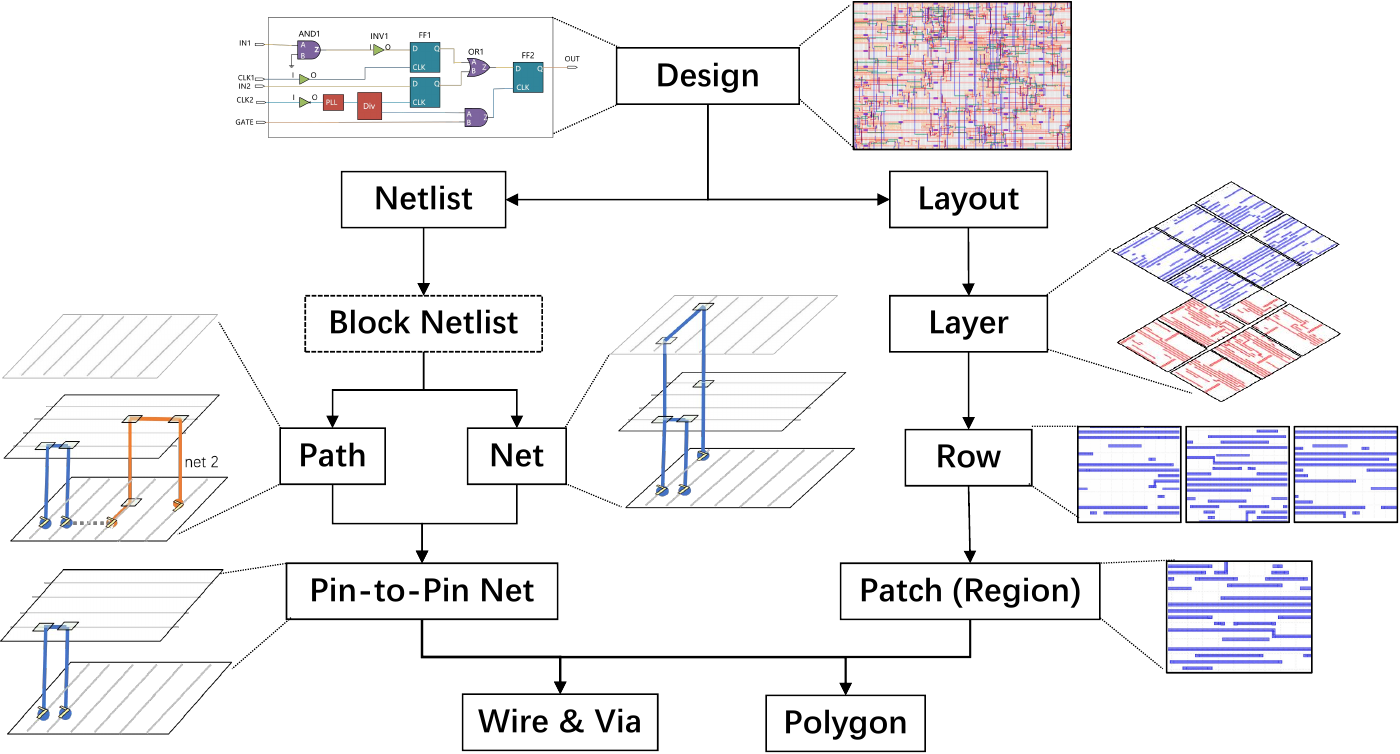}
\caption{Design-to-vector for chip data representation.}
\label{fig:vectorHierarchy}
\end{figure}

\subsubsection{Vectorization}
\label{subsec:vector}

Fig.~\ref{fig:vectorHierarchy} illustrates a hierarchical \textbf{design-to-vector} decomposition methodology for chip data representation. The methodology decomposes the overall design into two fundamental components: netlist and layout. The netlist captures logical connectivity between circuit cells, while the layout represents physical geometric information. Decomposition continues on both branches with increasing granularity. The netlist decomposes into path-level representations (encoding routing connectivity and signal propagation paths), and net-level representations (capturing individual point-to-point connection between specific pins). {Recognizing that path-level representations can introduce redundancy due to overlapping segments, we also provide a holistic graph-level representation. This allows users to directly access global connectivity information in a complete and non-redundant format, serving as a macro-level counterpoint to the fine-grained path data.} The layout decomposes into layer-level (manufacturing layers such as metal and via layers) and patch-level (localized spatial regions) data. Both branches converge at the geometric primitive level, encompassing three fundamental elements: wires (linear conductors), vias (inter-layer connection), and polygons (geometric shapes). 

\idel{The vectorized format provides key advantages for AI applications: it ensures computational efficiency through structured data organization while preserving the majority of original design information to maintain data integrity. This methodology transforms complex chip designs into structured vector formats inherently compatible with AI processing frameworks, enabling seamless integration across multiple abstraction levels. The resulting unified, fine-grained vectorized representation facilitates consistent data processing throughout the design hierarchy and supports diverse downstream applications.}

{This collection of structured, multi-level data constitutes what we term \textbf{Foundation Data}, pivotal for AI applications. It converts complex chip designs into a computationally efficient, structured representation that preserves high data integrity. Its inherent compatibility with AI processing frameworks enables seamless integration across all design abstraction levels. Ultimately, our vectorization methodology yields a unified, fine-grained data foundation that ensures consistent processing and supports diverse downstream applications.}

In AiEDA, vectorization is streamlined through simple and intuitive interfaces, as demonstrated in Listing~\ref{lst:vectorize_interface}. The framework supports flexible granularity control, allowing users to generate vectors at different abstraction levels based on specific application requirements.


\begin{figure}[htb]
\small
\begin{lstlisting}[style=pythonstyle, caption={{Unified Python API usage for vectorization.}}, label=lst:vectorize_interface]
from aieda.workspace import workspace_create
from aieda.data import RunVectors

def vectorize_data(ws_dir, tool, level): 
    ws = workspace_create(ws_dir, tool)
    vectors = RunVectors(ws)
    vectors.read_def(ws.input_def)
    match level: 
        case "net": 
            vectors.generateNet(ws.vec_dir)
        case "graph": 
            vectors.generateGraph(ws.vec_dir)
        case "path": 
            vectors.generatePath(ws.vec_dir)
        case "patch": 
            vectors.generatePatch(ws.vec_dir)
        case _: 
            feat.generateVectors(ws.vec_dir)
\end{lstlisting}
\end{figure}

\vspace{-5mm}
\subsection{Downstream Application} 
\label{subsec:downstream_application}

\begin{figure}
    \centering
    \includegraphics[width=0.99\linewidth]{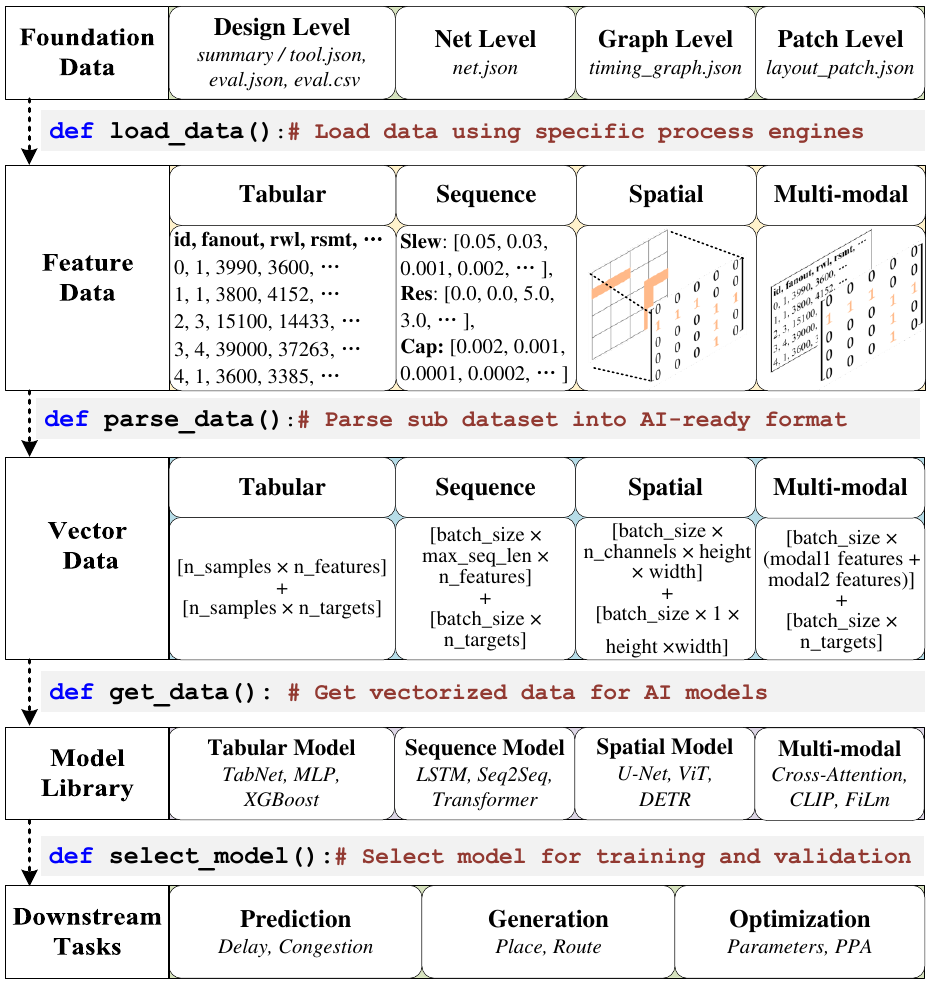}
    \caption{{Data processing pipeline for downstream tasks. The data structures depicted are illustrative examples and not an exhaustive list.}}
    \label{fig:downstreamVis}
\end{figure}

We develop \textbf{process engines} that {selectively} extract and organize data from the {Foundation Data}, {streamlining} preparation for downstream tasks. This workflow is shown in Fig.~\ref{fig:downstreamVis}. {Note that the data structures depicted in \cref{fig:downstreamVis} are illustrative examples and not an exhaustive list.}

We categorize process engines into two types: \textbf{specific} and \textbf{general} engines. Specific engines comprise {six} types: {design, net, graph, path, patch, and combined engines.} These engines follow a consistent three-stage approach: (1) loading the {Foundation Data} to filter task-specific {feature data} using \texttt{load\_data()}, (2) performing feature engineering and reorganizing the data into AI-ready formats (e.g., tabular, sequence, spatial) via \texttt{parse\_data()}, and (3) providing {vector} data to AI models through \texttt{get\_data()}. The combined  engine, specifically, merges data from {multi-level} representations (e.g., net-level and patch-level) to form multi-modal representations. We also provide a general visual engine for analyzing dataset characteristics.

{A key challenge in preparing this data is handling the variable dimensions inherent in different circuits and tasks (e.g., varying net counts or path lengths). Our process engines resolve this by applying modality-specific techniques: sequential data is padded or truncated to a uniform length, spatial data is organized into fixed-size grids, and graph data is managed by native batching mechanisms (e.g., PyG). This ensures the data from \texttt{get\_data()} is seamlessly compatible with mainstream AI frameworks, abstracting complex preprocessing from the end-user.
}

Furthermore, AiEDA supports loading diverse model libraries through \texttt{select\_model()}. For each data modality, we provide multiple models for training and validation in a unified environment. The model library is extensible, accommodating both third-party libraries and user-defined custom models encapsulated as classes. {To simplify development, AiEDA promotes a standardized ``config → process → model → trainer" workflow. This pattern, inspired by leading AI frameworks, lowers the barrier for contributing new models by providing clear, established precedents.}

By introducing process engines and model libraries, AiEDA standardizes the AAD pipeline from data loading to model validation, thereby advancing AAD methodology development.

\section{\textbf{iDATA}: Dataset for AI-Aided Design}
\label{sec:dataset}

\subsection{Data Source and Statistics}
\label{sec:dataset:source}

\begin{table*}[htbp]
\centering
\caption{Statistical characteristics of ``iDATA".}
\setlength{\tabcolsep}{10.5pt}
\label{tab:data}
\begin{tabular}{l rr r rrr rrr}
\toprule
\multirow{2}{*}{Circuits} &  \multicolumn{3}{c}{Design} & \multicolumn{2}{c}{Net} & \multicolumn{2}{c}{Path} & \multicolumn{2}{c}{Patch} \\
\cmidrule(lr){2-4} \cmidrule(lr){5-6} \cmidrule(lr){7-8} \cmidrule(lr){9-10}
 & \#Cells & \#Nets & \#Wires & \#Files & Size & \#Files & Size & \#Files & Size \\
\midrule
\texttt{s713} & 135 & 125  & 1426 & 121 &890K & 56 & 676K & 324 & 1.59M \\
\texttt{s44} & 178 & 128  & 2095 & 128 & 1.31M & 128 & 558K & 441 & 2.31M \\
\texttt{apb4\_rng} & 195 & 204  & 2230 & 169 & 1.39M & 132 & 1.29M & 441 & 2.45M \\
\texttt{gcd} & 297 & 270  & 3733 & 270 & 2.32M & 136 & 3.81M & 625 & 3.85M \\
\texttt{s1238} & 349 & 290  & 4998 & 290 & 3.00M & 24 & 284K & 576 & 4.27M \\
\texttt{s1488} & 380 & 325  & 6422 & 325 & 3.82M & 24 & 584K & 506 & 4.99M \\
\texttt{apb4\_arch} & 392 & 381  & 5122 & 346 & 3.19M & 304 & 1.45M & 841 & 5.14M \\
\texttt{apb4\_ps2} & 515 & 497  & 6542 & 432 & 4.04M & 372 & 5.72M & 1024 & 6.35M \\
\texttt{s9234} & 657 & 585  & 8592 & 577 & 5.24M & 456 & 5.36M & 1225 & 8.05M \\
\texttt{apb4\_timer} & 721 & 689  & 8960 & 653 & 5.65M & 332 & 4.51M & 1521 & 9.32M \\
\texttt{s13207} & 727 & 647 & 8182 & 636 &5.07M  &744  & 5.03M & 1600 & 8.83M \\
\texttt{apb4\_i2c} & 790 &727   &10248  &676  &6.37M  &568  &9.10M  &1521  &9.85M  \\
\texttt{s5378} &881  &774   &12422  &774  &7.62M  &564  &6.60M  &1600  &11.61M  \\
\texttt{apb4\_pwm} &974  &889   & 13596 &838  &8.36M  &480  &4.66M  &1936  &13.02M  \\
\midrule
\texttt{apb4\_wdg} &1029  &945   &13889  &910  &8.58M  &456  &7.69M  &1936  & 13.17M  \\
\texttt{apb4\_clint} & 1069  &1004   &13634  &969  &8.43M  &524  & 7.63M &1936  &13.07M  \\
\texttt{ASIC} &1228  &796   &10737  &796  &6.82M  &76  &660K  &121  &5.36M  \\
\texttt{s15850} &2088  &1926   &27941  &1925  &17.33M  &1724  &21.00M  &4225  &27.66M  \\
\texttt{apb4\_uart} & 5981 &5606  & 83268 &5555  &53.45M  &4652  &99.96M  &11449  &83.34M  \\
\texttt{s38417} &6028  &5573   &85054  &5573  &53.30M  &5764  &113.04M  &12544  &84.65M  \\
\texttt{s35932} & 6375 &5837   &81158  &5837  &52.02M  &6912  &33.66M  &14161  &93.73M  \\
\texttt{s38584} &7023  &6586   &97771  &6585  &61.51M  &4916  &62.76M  &12100  &94.02M  \\
\texttt{BM64} &9358  &9510   &132076  &9510  &85.63M  &5164  &183.90M  &16641  &142.47M  \\
\texttt{picorv32} &9430  &9077   &136455  &9010  &87.75M  &6560  &177.09M  &21316  &150.30M  \\
\texttt{PPU} & 9547 &8895   &140136  &8895  &90.05M  &9552  &164.43M  &20164  &143.55M  \\
\midrule
\texttt{blabla} &15154  &15672   &216427  &15671  &144.32M  &4396  &230.33M  &24649  &252.01M  \\
\texttt{aes\_core} &17940  &17371    &310215  &17371  &202.32M  &9876  &623.39M  &32041  &311.46M  \\
\texttt{aes} & 19181  &18117   &325550  &18117  &212.03M  &11940  &623.84M  &36100  &323.91M  \\
\texttt{salsa20} &21270  &20432   &291172  &20432  &190.66M  &14916  &439.55M  &40401  &320.98M  \\
\texttt{jpeg} &27671  &29160  &366397  &29160  &245.32M  &18128  &786.71M  &66049  &424.45M  \\
\texttt{eth\_top} & 42279 &38552    &646875  &38552  &434.20M  &20000  &562.18M  &169744  &967.16M  \\
\texttt{yadan} & 63514  &31280   &483369  &31280  &331.30M  &19832  &816.37M  &15376  &313.25M  \\
\midrule
\texttt{beihai} & 211236  &133086   &2161829  & 132424 & 1.53G & 52628 &4.87G  & 92256 &1.60G  \\
\texttt{SHMS} & 268721  &251772   &3610024  & 251686  & 2.51G & 70672 & 6.45G & 19153 &1.98G  \\
\texttt{nvdla} & 289344  &226974  &3708427  & 226904  & 2.69G & 20000 &1.21G  &28224  &2.40G   \\
\texttt{ZJUC} & 348985  &323195  &4880442 & 323175 &3.35G  &184416  &12.61G  &23968  &2.66G   \\
\texttt{iEDA23} & 368147  &335112   &5132004  &335026  & 3.50G &219852  & 21.55G & 28224 & 2.84G \\
\texttt{ysyx4S2} & 494962  &449847   &6967779   & 448950  &4.79G  &11864  &1.58G  &28224  &3.84G \\
\texttt{beihai2} & 582645  &393308   &5491501  & 391062 & 4.21G &70264  &11.04G  & 126630 &4.19G\\
\texttt{AIMP} & 742210  &535618    &9980714   &534081  &8.07G  &26296  &1.48G  &77841  &8.08G\\
\texttt{AIMP2} & 816677  &560525    &9133333   &558606  &7.03G  &22540  &3.81G  &134246  &7.24G  \\
\midrule
\texttt{ysyx6} & 1090820  &1029515   &15486068   &1029496  &10.98G  &184700  &26.41G  &60501  &8.95G \\
\texttt{wukong} & 1102663  &1032718    &15653639   &1031971  & 10.98G &20120  &1.88G  &39480  &9.05G \\
\texttt{ysyx41} & 1123368  &1105564    &18248292   &1105478  &12.24G  &219848  &21.67G  &28224  &9.73G \\
\texttt{ysyx42} & 1173610  &1147953    &17416249   &1147867  &11.98G  &199704  &14.18G  & 36100 &9.56G \\
\texttt{T1} & 1262053  &1227098    &18769036   & 1227019 &12.63G  &40000  &2.62G  &24649  &10.03G \\
\texttt{T1\_mach} & 2222669  &2162147    &33115508   &2162068  &22.82G  &40000  &3.08G  &43681  &18.27G\\
\texttt{nanhu-G} & 2793215  &2646672    &42524007   & 2643701 & 27.84G  &20136  &1.70G  &75040  &25.04G \\
\texttt{openC910} & 3282828  &2948743   &52259408  & 2942510 & 36.25G & 40588 & 2.85G & 152100 &33.82G  \\
\texttt{T1\_sand} & 4816399  &4728816    &79050737   &4728737  &50.22G  &40000  &5.99G  &77841  &44.13G \\
\midrule
Total & 23.26M & 21.47M  & 347.15M & 21.45M & \textbf{235.91G} & 1.63M & \textbf{149.87G} & 1.61M & \textbf{207.18G} \\
\bottomrule
\end{tabular}
\end{table*}

{This section introduces iDATA, a large-scale, open-source dataset built to facilitate AAD research. The complete dataset, derived from 50 real-world chip designs, totals approximately 600 GB of structured Foundation Data, excluding the original raw design files.}~\cref{tab:data} summarizes the statistical characteristics of our dataset comprising 50 designs. The dataset encompasses diverse chip designs including digital signal/image processors (DSP/ISP), peripheral/interface circuits, functional modules, memories, CPUs, GPUs, and SoCs, primarily sourced from open repositories (OSCPU\cite{OpenSourceChip}, OSCC\cite{OSCCIPProject}, OpenLane\cite{EfablessOpenlane2cidesignsContinuous}, ISCAS89\cite{IsprasHdlbenchmarksCollection}, CHIPS Alliance\cite{CHIPSAlliance}) and internal projects. All designs undergo RTL synthesis in 28nm technology, followed by complete physical design using \textbf{Innovus} and data extraction via \textbf{iEDA}. {We chose Innovus for the physical design process because results from commercial tools are widely regarded as a high-quality ``ground truth" in the AI research community, providing a stable and reliable benchmark.} {The iDATA dataset is the direct artifact generated from this workflow using our AiEDA library.} Since routing results contain comprehensive physical design information, we construct our dataset based on routing outcomes, corresponding to design-level, net-level, {graph-level}, path-level, and patch-level vectorizations. {Moreover, to support cross-stage predictive tasks, we have explicitly preserved key placement-stage features (e.g., cell density) within the Foundation Data.}

\textbf{Design-level} vectorization captures basic statistics, tool metrics, and evaluation metrics. \cref{tab:data} presents a subset of basic statistical data, specifically cell, net, and wire counts. Cell counts range from 135 to 4,816,399; net counts from 125 to 4,728,816; and wire counts from 1,426 to 79,050,737. This broad spectrum demonstrates the extensive design scale coverage in our dataset. 

Table~\ref{tab:data} also provides statistics on file counts and storage sizes across net-level, path-level, and patch-level representations. For \textbf{net-level} data, \textit{\#Files} may be fewer than \textit{\#Nets} since files are not generated for nets without wires. For \textbf{path-level} data, we extract only paths containing driver pins, potentially resulting in significantly fewer files than \textit{\#Nets}. \idel{Additionally, }Designs with exceptionally large path counts utilize configurable parameters to limit path generation. {To ensure data completeness, we generate a comprehensive \textbf{graph-level} representation for each netlist, capturing its full topology.} For \textbf{patch-level} data, patch sizes are adjustable to accommodate different design scales. For small circuits, the default patch size is set to 9 times the pitch width, while for large designs, it is 180 times the pitch width. This configurability enables flexible dataset generation tailored to specific design requirements.

To assess computational requirements and scalability, vectorization was performed using 32 threads on an Intel Xeon Platinum 8268 CPU with 1.5TB RAM. 
Vectorization time ranges from tens of minutes to hours, depending on design complexity and parameter settings (e.g., patch size). The 1.5TB memory configuration proved sufficient even for our largest design (\texttt{T1\_sand}), demonstrating scalability for incremental dataset expansion.

\subsection{{Foundation Data Generation}}
\label{sec:source data}

\begin{figure}[t]
    \centering
    \subfigure[]{
        \includegraphics[width=0.99\linewidth]{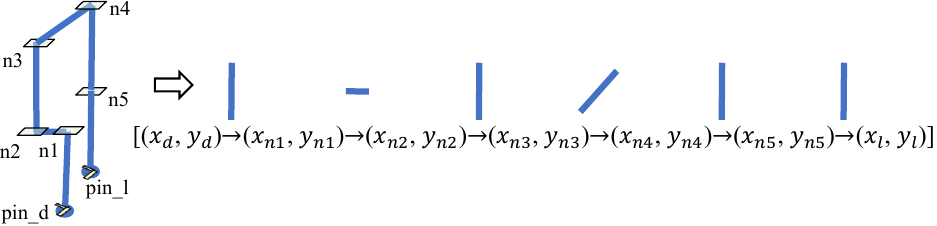}
    }\label{fig:source-data_a}
    \hfill
    \subfigure[]{
        \includegraphics[width=0.99\linewidth]{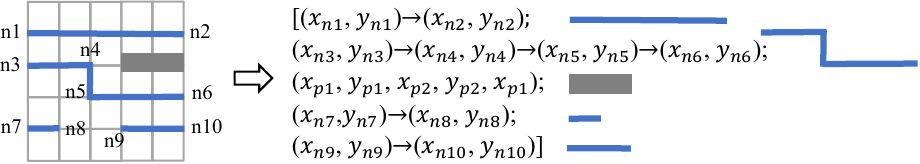}
    }\label{fig:source-data_b}
    \caption{{Foundation Data generation. (a) Pin-to-pin net. (b) Patch.}}
    \label{fig:source-data}
\end{figure}

{\cref{fig:source-data} illustrates how Foundation Data is generated at the net, path, and patch levels.} In~\cref{fig:source-data}(a), a pin-to-pin net connects driver pin \textit{pin\_d} to load pin \textit{pin\_l} through intermediate nodes \textit{n1}-\textit{n5}. This net is decomposed into six wire segments, which can be \idel{easily} represented as a vector, as shown in~\cref{fig:source-data}(a). Following the decomposition approach described in~\cref{fig:vectorHierarchy}, both paths and nets can be broken down into multiple pin-to-pin nets. Therefore, source paths and nets can be readily vectorized in the same manner as in~\cref{fig:source-data}(a). Additionally, \cref{fig:source-data}(b) shows a patch consisting of four wire segments and an obstacle polygon, all stored in vector form. The vector includes the $x$ and $y$ coordinates of the connected nodes, along with the contour coordinates of the obstacle polygon.

\begin{figure}[!t]
    \centering
    \includegraphics[width=1\linewidth]{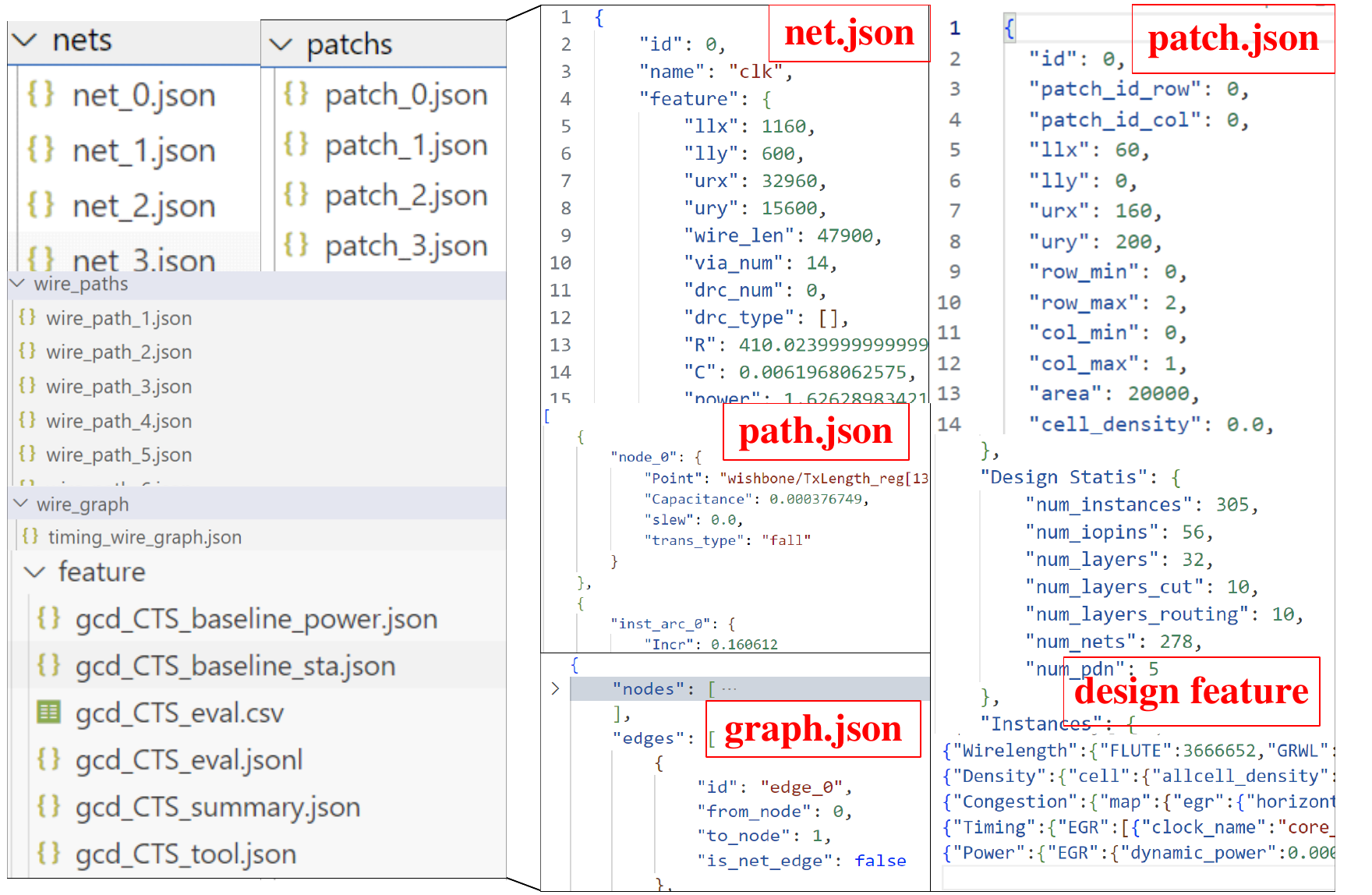}
    \caption{File organization structure of the vectorized dataset.}
    \label{fig:vectorFile}
\end{figure}

{We generate the Foundation Data using the method in \cref{fig:source-data} and store it in files as shown in \cref{fig:vectorFile}.} \idel{We use the manner in~\cref{fig:source-data} to \idel{vectorize}{generate} the \idel{source data}{Foundation Data}, and store them into some files as in~\cref{fig:vectorFile}. }\idel{In addition to vectorized source data, the data files in \cref{fig:vectorFile} also include rich feature information.}\cref{tab:data_levels} presents the key data at each hierarchical level. \textbf{Design} level provides basic statistics, tool metrics, and evaluation metrics with outputs ranging from scalar values to spatial maps. \textbf{Net} level stores individual net information including net features, parasitic parameters, and wire geometry details. {\textbf{Graph} level represents the netlist as a topological graph, defining nodes and edges to capture the full circuit structure.} \textbf{Path} level captures timing path data with detailed node properties, signal parameters, and decomposed interconnect information for each wire segment. \textbf{Patch} level employs uniform spatial division to generate patches containing position identifiers, quality indicators, and layer-specific features for localized analysis.

\begin{table}[!t]
\centering
\caption{{Data content vectorized at each hierarchical levels.}}
\label{tab:data_levels}
\setlength{\tabcolsep}{8.5pt}
\begin{tabular}{|p{0.7cm}|p{6.8cm}|}
\hline
\textbf{Level} & \textbf{Key Data} \\
\hline
\textbf{Design} & 
Layout dimensions, area utilization, net/cell counts and types, routing layers, pin distribution, tool runtime metrics (e.g., \textit{``buffer\_insertion"} during CTS), wirelength, density, timing, power, congestion maps\\
\hline
\textbf{Net} & 
Net name/features, pin information, bounding box, parasitic R/C, delay, slew, power, fanout, DRC counts/types, wire segments and geometric properties \\
\hline
\textbf{Graph} & 
Node definitions (unique ID, name), pin/port identifiers, and edge definitions specifying connectivity\\
\hline
\textbf{Path} & 
Node positions, capacitance, slew rates, transition types, cell/interconnect delays, edge decomposition, wire R/C parameters, input/output slew, wire delay parameters \\
\hline
\textbf{Patch} & 
Patch ID, position identifiers, boundaries, grid indices, density metrics, RUDY/EGR congestion, timing, power, IR drop, sub-net decomposition, layer-wise wire features \\
\hline
\end{tabular}
\end{table}
\vspace{-2mm}

\begin{table}[!t]
\centering
\caption{{Net-level fidelity comparison for the \texttt{s713} design.}}
\label{tab:net_fidelity}
\setlength{\tabcolsep}{8.5pt}
\begin{tabular}{lrrr}
\toprule
\textbf{Metric} & \textbf{Original} & \textbf{Reconstructed} & \textbf{Fidelity Ratio} \\
\midrule
WNS (ns) & -0.972 & -0.989 & 0.983 \\
TNS (ns) & -3.887 & -3.987 & 0.975 \\
Violating Paths & 7 & 7 & 1.000 \\
Total Power (W) & 0.0621 & 0.0622 & 0.998 \\
\bottomrule
\end{tabular}
\end{table}

\subsection{{Fidelity and Accuracy Loss Validation}}
\label{sec:fidelity}
\begin{figure}[!t]
\centering
\includegraphics[width=0.9\linewidth]{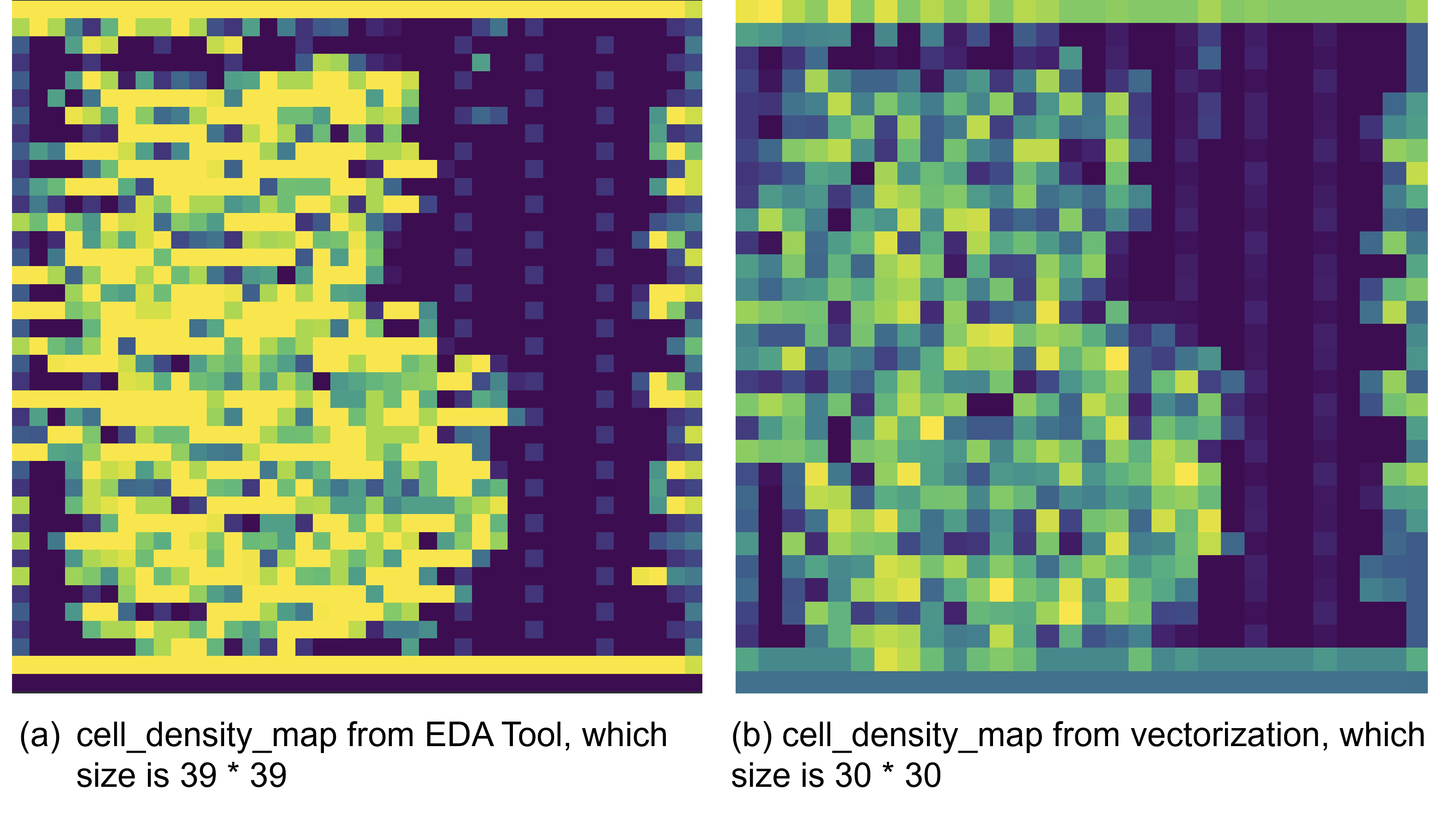} 
\caption{{Patch-level fidelity comparison for the \texttt{s713} design.}}
\label{fig:patch_fidelity}
\end{figure}

{The design-to-vector paradigm presents a controllable trade-off between data fidelity and processing efficiency. This accuracy loss, primarily stemming from discretization techniques like layout gridding, is minimal and considered acceptable for AI-EDA tasks that prioritize spatial patterns over absolute coordinate precision. The process's high fidelity is illustrated with the \texttt{s713} designs.

At the \textbf{net-level} (logical topology), we reconstructed a DEF file from its Foundation Data (\texttt{net.json}) and compared it to the original. The results, summarized in \cref{tab:net_fidelity}, revealed negligible deviation. Key metrics such as the number of timing-violating paths remained identical, while fidelity ratios for timing (WNS, TNS) and power approached 1.0, confirming the preservation of critical electrical characteristics. At the \textbf{patch-level} (geometric layout), a comparison between the original and the reconstructed cell density maps is shown in \cref{fig:patch_fidelity}. Despite a reduction in resolution, the reconstructed map clearly preserved the original's key spatial features. This visual similarity was quantified by a high correlation coefficient of 0.993. These results confirm our process preserves critical electrical characteristics and spatial distributions, making the minimal accuracy loss an acceptable trade-off for the significant gains in research efficiency and accessibility.

}

\subsection{Data Insight Analysis}
\label{subsec:data analysis}

Leveraging our process engines (Fig.~\ref{fig:downstreamVis}), we efficiently analyze dataset characteristics across multiple granularities. This analysis has a twofold objective: to provide a quantitative overview of the iDATA dataset and to establish a foundation for uncovering novel EDA insights. By creating a statistical baseline and validating fundamental domain principles, we enhance the dataset's accessibility and reliability for both AI researchers and EDA experts. Accordingly, our statistical analysis focuses on the feature-rich Design, Net, Path, and Patch levels. The Graph-level data is treated separately; as a pure topological representation (nodes and edges) intended for direct use in graph-centric AI models like GNNs, its value is structural rather than statistical. Therefore, it is excluded from the following characteristic summary.

\begin{figure}[htb]
\small
\begin{lstlisting}[style=pythonstyle, caption={Loading data from iDATA.}, label=lst:data_load]
from aieda.workspace import workspace_create
from aieda.data import DataVectors

def vectors_load(workspace): 
    data = DataVectors(workspace)
    
    layers = data.load_layers()
    vias = data.load_vias()
    cells = data.load_cells()
    instances = data.load_instances()
    nets = data.load_nets()
    patchs = data.load_patchs()
    instance_graph = data.load_instance_graph()
    timing_graph = data.load_timing_graph()
    timint_paths = data.load_timing_paths()
\end{lstlisting}
\end{figure}

\begin{figure}[!t]
    \centering
    \subfigure[]{
        \includegraphics[width=0.43\linewidth]{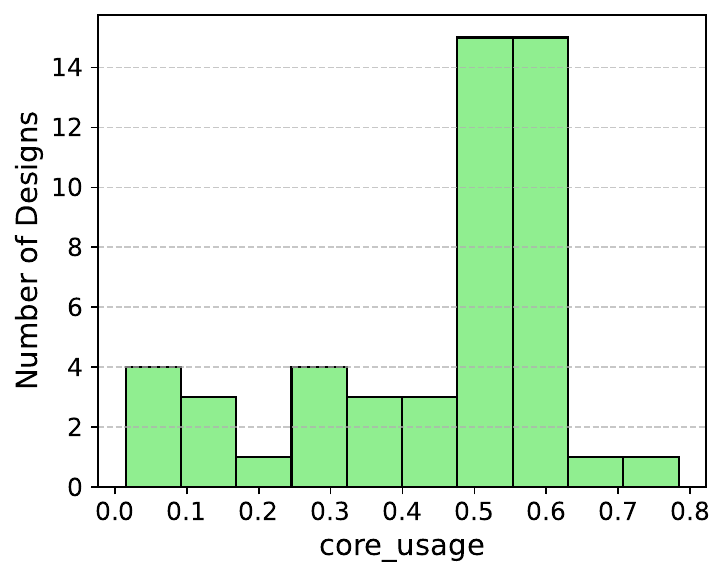}
    }
    \hfill
    \subfigure[]{
        \includegraphics[width=0.43\linewidth]{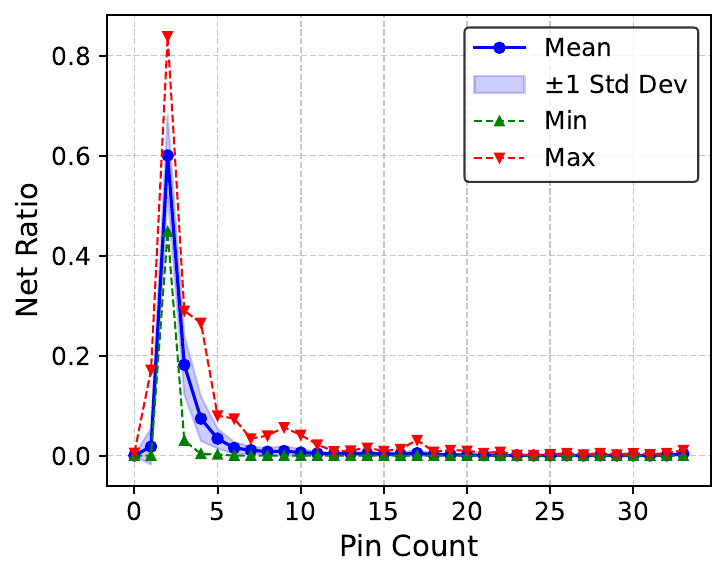}
    }
    \hfill
    \subfigure[]{
        \includegraphics[width=0.43\linewidth]{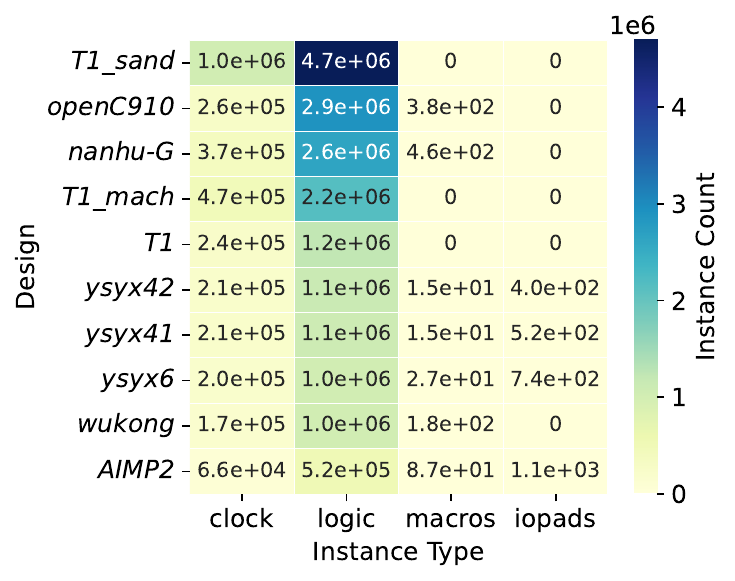}
    }
    \hfill
    \subfigure[]{
        \includegraphics[width=0.43\linewidth]{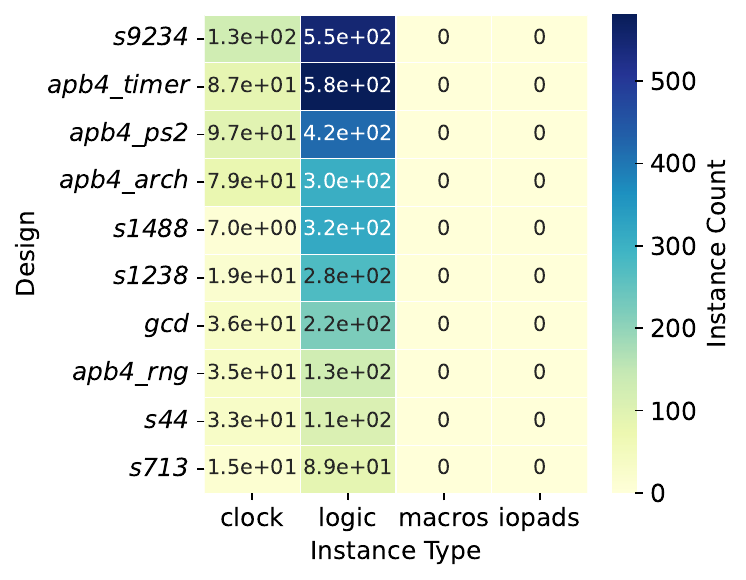}
    }
    \caption{Design-level characteristics. (a) Core usage. (b) Pin count. (c)-(d) Instance type distribution.}
    \label{fig:design_characteristics}
\end{figure}

\begin{figure}[t]
    \centering
    \subfigure[]{
        \includegraphics[width=0.43\linewidth]{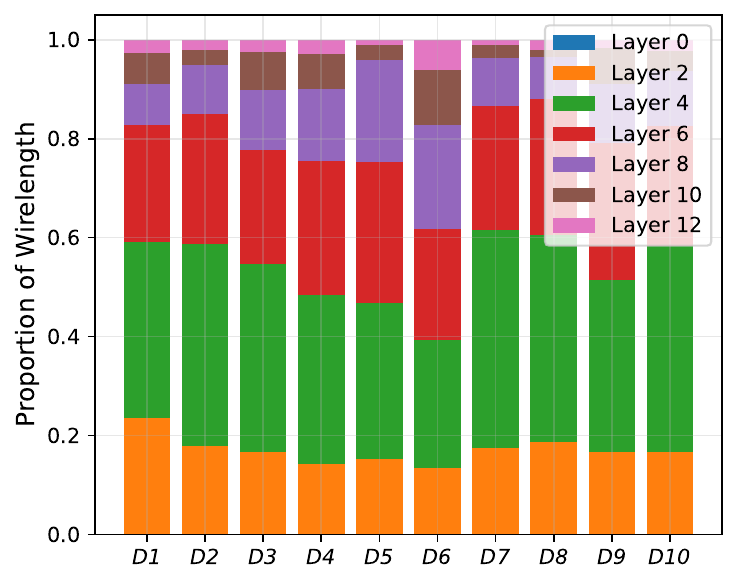}
    }
    \hfill
    \subfigure[]{
        \includegraphics[width=0.43\linewidth]{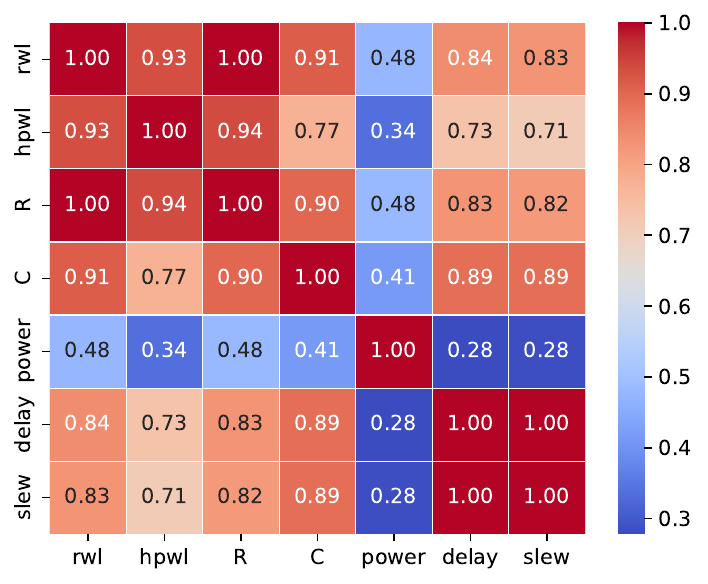}
    }
    \caption{Net-level characteristics. (a) Wirelength distribution across layers. (b) Correlation matrix of physical metrics (RWL, HPWL) and electrical parameters (R, C, power, delay, slew).}
    \label{fig:net_character}
\end{figure}

\textbf{Design-Level Characteristics.} We examine three key metrics: core usage, pin count distribution, and instance type composition. Core usage represents the ratio of total instance area to core area. Fig.~\ref{fig:design_characteristics}(a) shows that over half the designs exhibit core usage within 0.5-0.6. Pin count distribution analysis in Fig.~\ref{fig:design_characteristics}(b) reveals that nets with 2-3 pins predominate, accounting for approximately 80\% of all nets. The shaded regions represent $\pm$1 standard deviation across different pin counts. Instance types are categorized into four classes: clock, logic, macro, and IO pad. Fig.~\ref{fig:design_characteristics}(c) and (d) present instance type distributions for the top 10 and bottom 10 designs by total instance count, respectively. The analysis shows clock instances constitute 14.93\% on average, logic instances 85.04\%, macro instances 0.01\%, and IO pads 0.02\%.

\textbf{Net-Level Characteristics.} We analyze both physical metrics (RWL, HPWL) and electrical parameters (R, C, dynamic power, average delay, and delta slew). Physical metrics characterize routing geometry, while electrical parameters reflect performance and timing characteristics. Fig.~\ref{fig:net_character}(a) illustrates wirelength distribution across layers for various designs. Odd-numbered layers represent vias, while even-numbered layers correspond to metal routing layers, with Layer 0 reserved for cell placement. The distribution shows that Layer 4 accommodates approximately 40\% of total wirelength, while Layers 2 and 6 each account for approximately 20\%. Wire utilization decreases with increasing layer numbers. Fig.~\ref{fig:net_character}(b) presents the correlation matrix, where the high correlation coefficient (0.93) between RWL and HPWL validates HPWL as an effective predictor for actual routing length, confirming the efficacy of HPWL optimization during placement. Both RWL and HPWL exhibit strong correlations with R and C, indicating that wirelength reduction effectively mitigates parasitic effects. The correlation analysis reveals that physical design optimization (reducing wirelength) leads to improved electrical performance through reduced parasitic C, which simultaneously enhances delay and signal transition quality.

\begin{figure}[!t]
    \centering
    \subfigure[]{
        \includegraphics[width=0.42\linewidth]{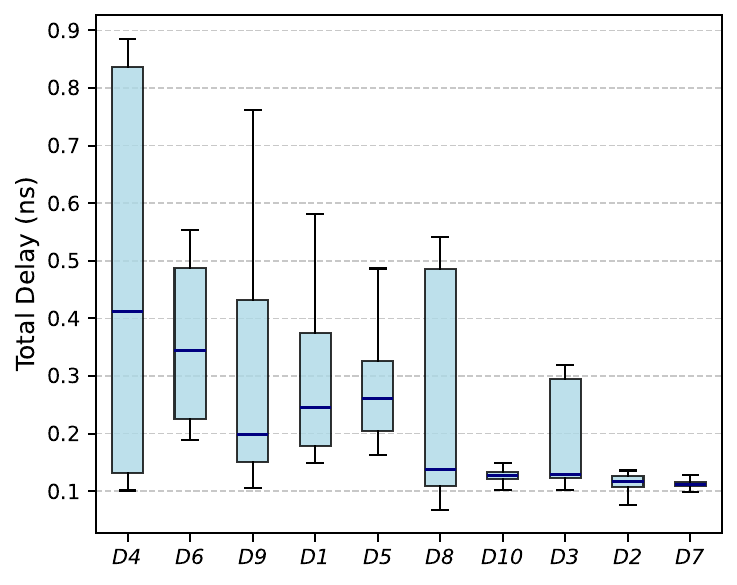}
    }
    \hfill
    \subfigure[]{
        \includegraphics[width=0.42\linewidth]{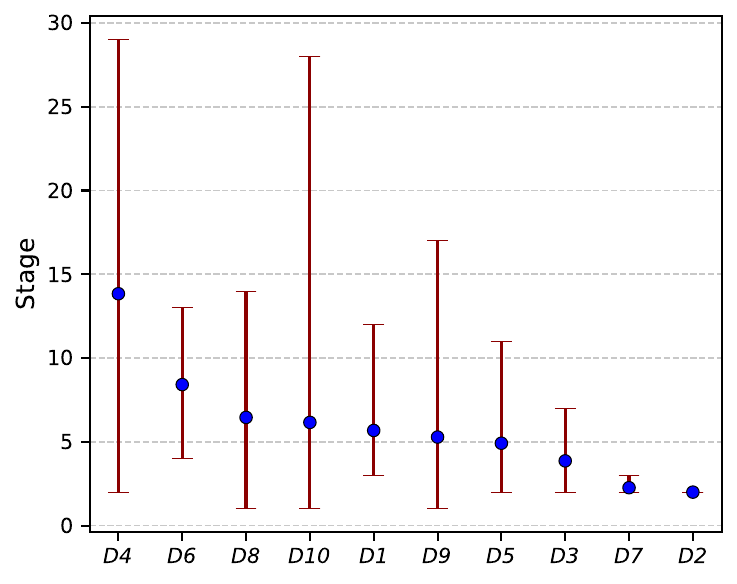}
    }
    \hfill
    \subfigure[]{
        \includegraphics[width=0.43\linewidth]{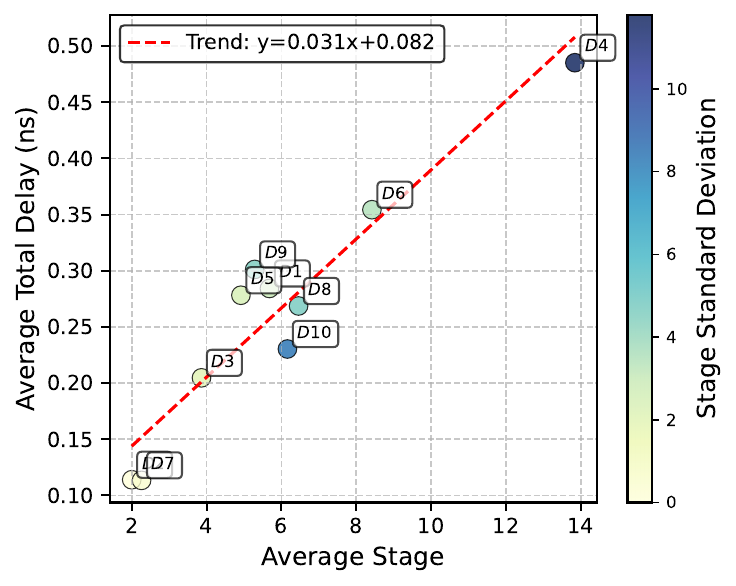}
    }
    \hfill
    \subfigure[]{
        \includegraphics[width=0.43\linewidth]{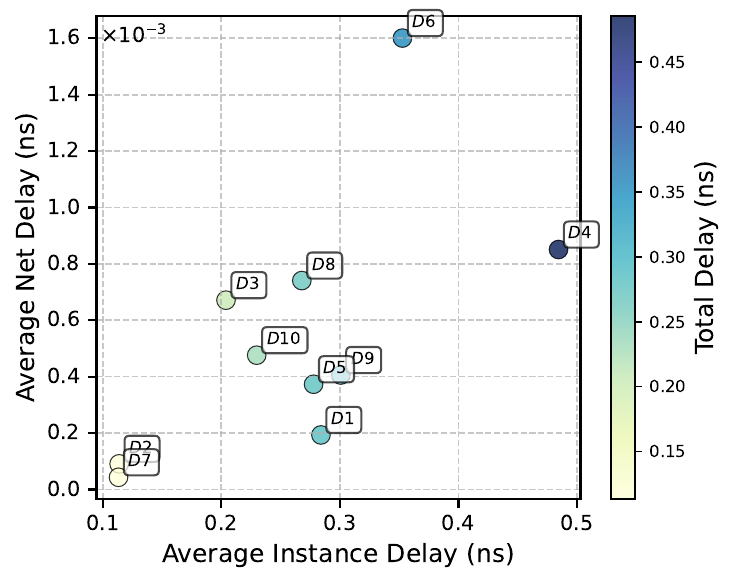}
    }
    \caption{Path-level characteristics. (a) Total delay distribution. (b) Stage count distribution. (c) Total delay vs. stage count correlation. (d) Instance vs. net delay  comparison.}
    \label{fig:path_character}
\end{figure}

\textbf{Path-Level Characteristics.} We analyze four critical timing metrics: instance delay, net delay, total delay, and stage count. Timing paths consist of instances interconnected by nets, with delays categorized as instance and net delays. Total delay accumulates instance and net delays along the path, while stage count indicates the number of instance-net pairs traversed. Fig.~\ref{fig:path_character}(a) and (b) present box plots of total delay and average stage count, revealing concentrated distributions within narrow ranges. Fig.~\ref{fig:path_character}(c) demonstrates an approximately linear relationship between total delay and average stage count, reflecting that increased stage count results in longer paths with higher cumulative delay. Fig.~\ref{fig:path_character}(d) shows the scatter plot of average instance delay versus average net delay, indicating both metrics distribute within specific ranges, with instance delay significantly exceeding net delay.

\begin{figure}[!t]
    \centering
    \includegraphics[width=1\linewidth]{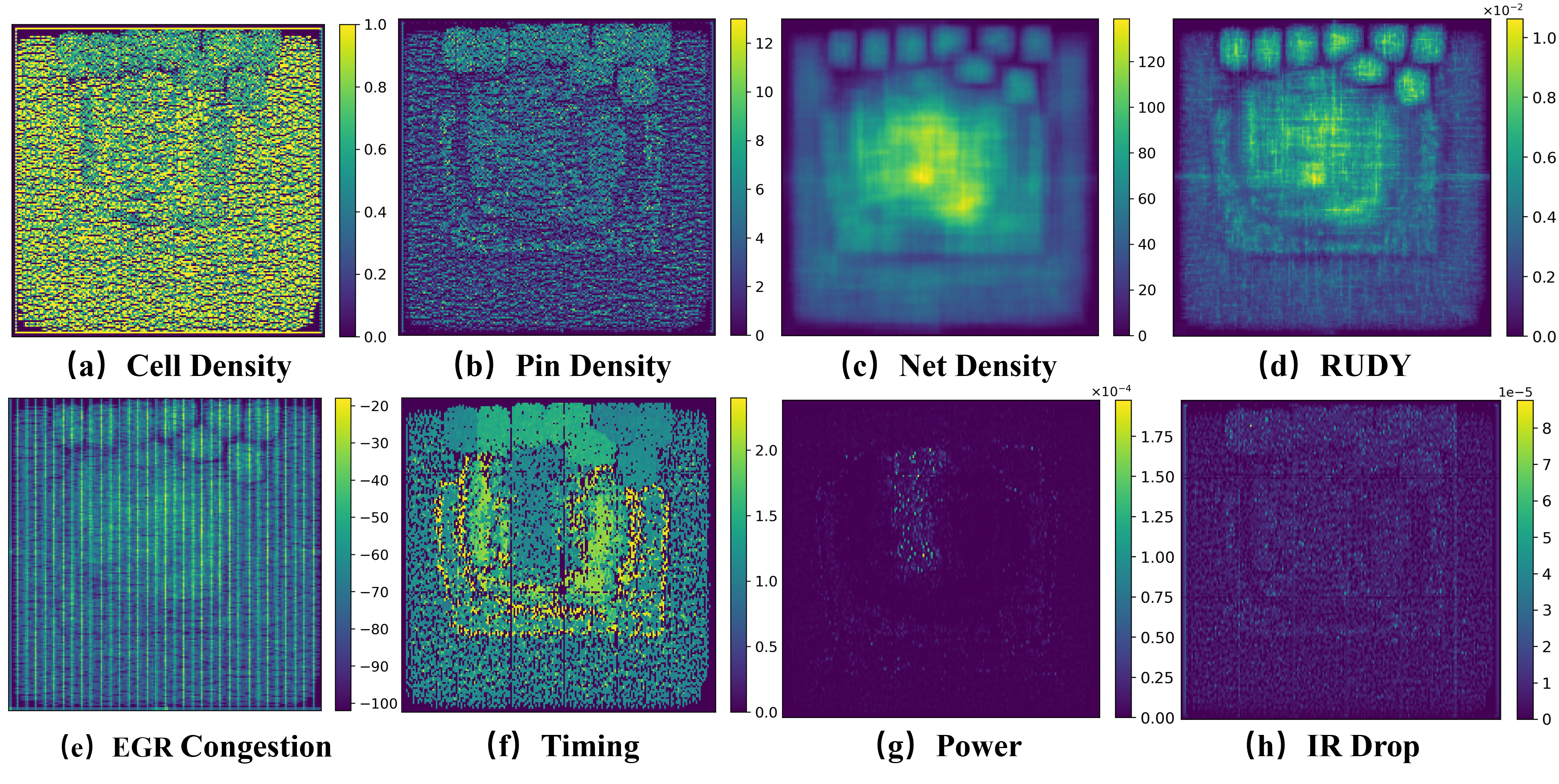}
    \caption{Spatial distribution of key features in \texttt{aes} chip layout.}
    \label{fig:patch_macro}
\end{figure}

\begin{figure}[!t]
    \centering
    \subfigure[Wire Density vs. Congestion]{
        \includegraphics[width=0.46\linewidth]{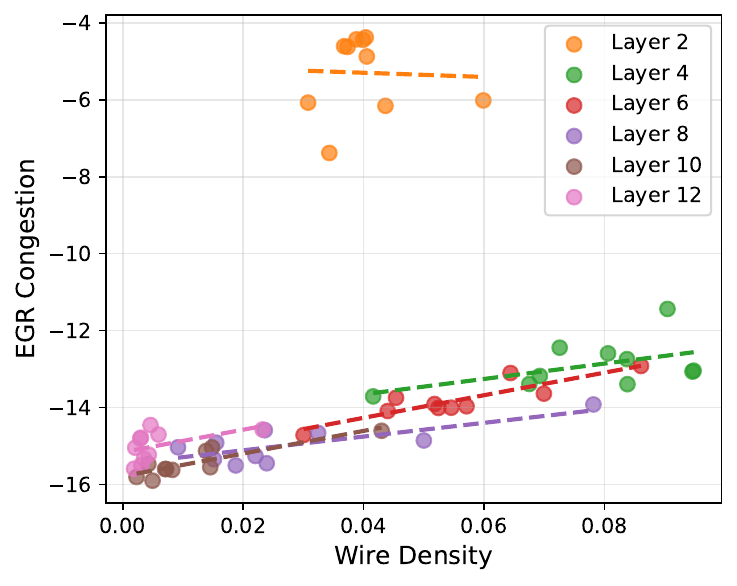}
    }
    \hfill
    \subfigure[Feature Correlation Analysis]{
        \includegraphics[width=0.46\linewidth]{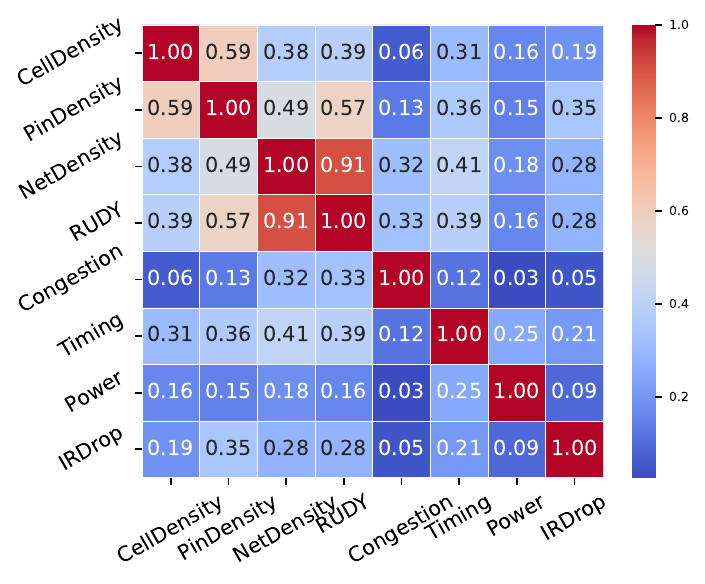}
    }
    \caption{Patch-level characteristics. (a) Regression analysis between wire density and congestion across metal layers. (b) Correlation matrix of patch-level features. }
    \label{fig:patch_character}
\end{figure}

\textbf{Patch-Level Characteristics.} We examine characteristics from both macro and micro perspectives. At the macro level, reassembled patches reconstruct complete design layouts. Fig.~\ref{fig:patch_macro}(a)-(h) visualize eight feature maps of the \texttt{aes} chip, enable layout-level tasks such as routability and IR drop prediction as demonstrated in CircuitNet\cite{chaiCircuitNetOpenSourceDataset2023b}. At the micro level, each patch represents an individual sample. Fig.~\ref{fig:patch_character}(a) demonstrates an approximately linear relationship between wire density and congestion across different layers, with congestion predominantly concentrated in Layer 2. Fig.~\ref{fig:patch_character}(b) presents correlation analysis of patch features. Notably, congestion shows moderate correlation with net density (0.32) but weak correlation with cell density (0.06), indicating that optimizing cell density alone is insufficient for congestion mitigation in routability-driven placement.

\begin{figure}[t]
    \centering
    \includegraphics[width=0.75\linewidth]{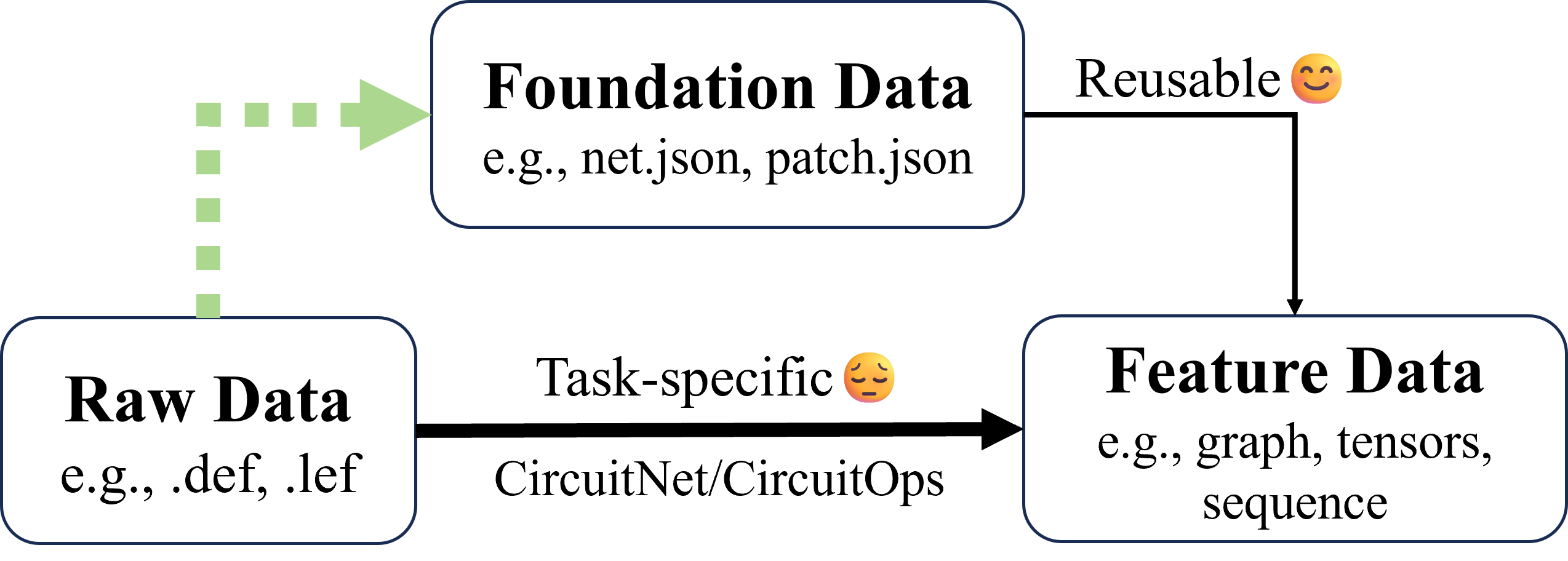}
    \caption{{The relationship between Raw Data, Foundation Data, and Feature Data, 
    and the associated development effort for new AI tasks. The thicker the line, the greater the develop effort.}}
    \label{fig:develop_effort_1}
\end{figure}

\cref{tab:dataset-comparison} compares existing open-source datasets for physical design across {three}\idel{four} dimensions: diversity, scale, {and }characteristics. \idel{, and applications.}While all these datasets provide raw data, they differ significantly in their focus and capabilities. CircuitNet\cite{chaiCircuitNetOpenSourceDataset2023b,jiangCircuitNet20Advanced2023} extracts layout and graph features for prediction tasks but sacrifices substantial data information. EDA-Schema~\cite{shresthaEDAschemaGraphDatamodel2024} specializes in metric extraction for prediction applications. ChipBench\cite{wang2024benchmarking} provides comprehensive raw data but lacks extensive feature data, being primarily designed for metric analysis rather than AI tasks. 
\idel{In comparison, iDATA originates from more diverse designs with larger scale, supports both feature and metrics extraction, and additionally extracts substantial source data through vectorization methods, making the dataset suitable for multiple types of tasks, including analysis, prediction, optimization, generation, and pre-training, etc.}
{In contrast, iDATA originates from more diverse and larger-scale designs and uniquely provides Foundation Data. This representation retains rich, original design information, setting it apart from the task-specific and often information-lossy Feature Data common to other approaches (as illustrated in \cref{fig:develop_effort_1}). The richness of this Foundation Data makes iDATA highly versatile and directly applicable to a wide range of tasks, including analysis, prediction, optimization, and generation. Consequently, our approach lowers the development effort for new AI applications by providing a more powerful and flexible starting point.}

\begin{table}[t]
\centering
\setlength{\tabcolsep}{4.5pt}
\caption{{Comparison of open-source datasets for physical design. Scale means the range of cell counts. All datasets provide raw data.} }
\begin{tabular}{c|c|c|c}
\toprule
Datasets & Chips & Scale & Characteristic     \\
\midrule
\makecell{CircuitNet \cite{chaiCircuitNetOpenSourceDataset2023b,jiangCircuitNet20Advanced2023}} &8 &46K-1.48M  & \makecell{Layout/Graph Features}  \\
\hline
\makecell{EDA-Schema \cite{shresthaEDAschemaGraphDatamodel2024}} &20 &0.47K-0.12M  & {Design Process Metrics}   \\
\hline
\makecell{ChipBench \cite{wang2024benchmarking}}  &20 &0.82K-0.86M & Raw Data Only \\
\hline
\makecell{\textbf{iDATA} \textbf{(This work)}}  & 50 &0.14K-4.82M &\makecell{{Foundation Data}}   \\
\bottomrule
\end{tabular}
\label{tab:dataset-comparison}
\end{table}

\section{AI-Aided Design Tasks and Results}
\label{sec:tasks}
A key advantage of the AiEDA library and the iDATA dataset is their native support for multi-modal and multi-source AI-EDA tasks. By preserving the unique identifiers and hierarchical relationships between different design objects (e.g., nets, patches, patches, and graph) during vectorization, AiEDA facilitates the fusion of data from logical, physical, and timing domains. This aligned data structure empowers researchers to easily implement sophisticated feature engineering and explore advanced model architectures. The applications in this section put these principles into practice, demonstrating tasks that leverage data from a single source as well as those that benefit from the fusion of multiple data modalities.

AiEDA's role is to serve as an enabling platform, providing standardized data and model interfaces for this closed loop. The specific optimization algorithms, as you noted, are naturally tightly coupled with the EDA tool's C++ implementation. To illustrate how AiEDA supports this complete workflow, we present a concrete example: using an AI model to predict post-routing wirelength during placement to guide detailed placement optimization.
The complete, iterative flow is as follows:

\begin{enumerate}
\item \textbf{Data Generation and Vectorization:}
Using AiEDA’s Flow and Data APIs, multiple design flows are executed. For each design, the vectorization engine extracts net-level features for wirelength prediction, forming a standardized training dataset.

\item \textbf{Model Training:}
Based on the vectorized data, a wirelength prediction model (e.g., TabNet) is trained to estimate post-routing wirelength from placement-stage features.

\item \textbf{Model Deployment and Integration:}
The trained model is exported to ONNX format and integrated into the open-source C++ EDA tool (iEDA) via ONNX Runtime. The \texttt{export\_to\_onnx()} function in our repository handles model export and enables real-time inference in iEDA (\url{https://github.com/OSCC-Project/AiEDA/blob/master/aieda/ai/net_wirelength_predict/tabnet_model.py}).

\item \textbf{AI-Guided Optimization:}
In iEDA’s detailed placement, the optimizer calls the ONNX model for real-time wirelength prediction instead of using traditional HPWL heuristics, achieving more accurate and adaptive optimization decisions.
\end{enumerate}

\begin{figure}[!t]
\centering
\includegraphics[width=1\linewidth]{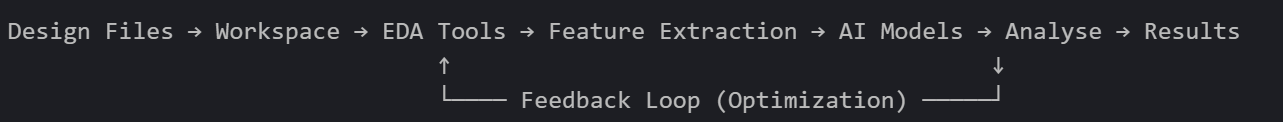}
\caption{The complete AiEDA data flow, from data generation and feature extraction to model training and feedback into the EDA tool for guided optimization.}
\label{fig:feedback_loop}
\end{figure}

While the engineering effort to integrate the model into the EDA tool's core is substantial, the fact that this entire flow operates smoothly is a direct testament to AiEDA's success as a\textbf{ robust bridge between AI and EDA}. The complete data flow is illustrated in  \cref{fig:feedback_loop}.

All experiments presented in this section are built upon this principle. They leverage iDATA's \textbf{Foundation Data} and AiEDA's \textbf{process engines} to load and prepare data for the models. This approach significantly reduces the development effort compared to other frameworks, such as CircuitNet or CircuitOps, where developing a new task would necessitate building a data processing pipeline from the complex raw data from scratch. The primary goal of these applications is to demonstrate the versatility of our framework through accessible \textbf{proof-of-concept} examples. They are intended to showcase a breadth of capabilities, rather than to establish new state-of-the-art results, which we view as important future work this framework now helps facilitate.

To ensure representativeness, our downstream tasks cover three AI categories (prediction, generation, optimization) across {five} vectorization levels (design, net, {graph}, path, patch).
This section details five representative downstream tasks to demonstrate the effectiveness of our AAD library and dataset, including methodologies and experimental results. All experiments were conducted on a system equipped with Intel Xeon Platinum 8380 CPU@2.30GHz (160 cores), 512 GB RAM, and NVIDIA A100 GPU (40 GB VRAM), running Ubuntu 18.04.5 with PyTorch 2.5.1 and CUDA 12.0.

\begin{figure}[t]
    \centering
    \subfigure[]{
        \includegraphics[width=0.45\linewidth]{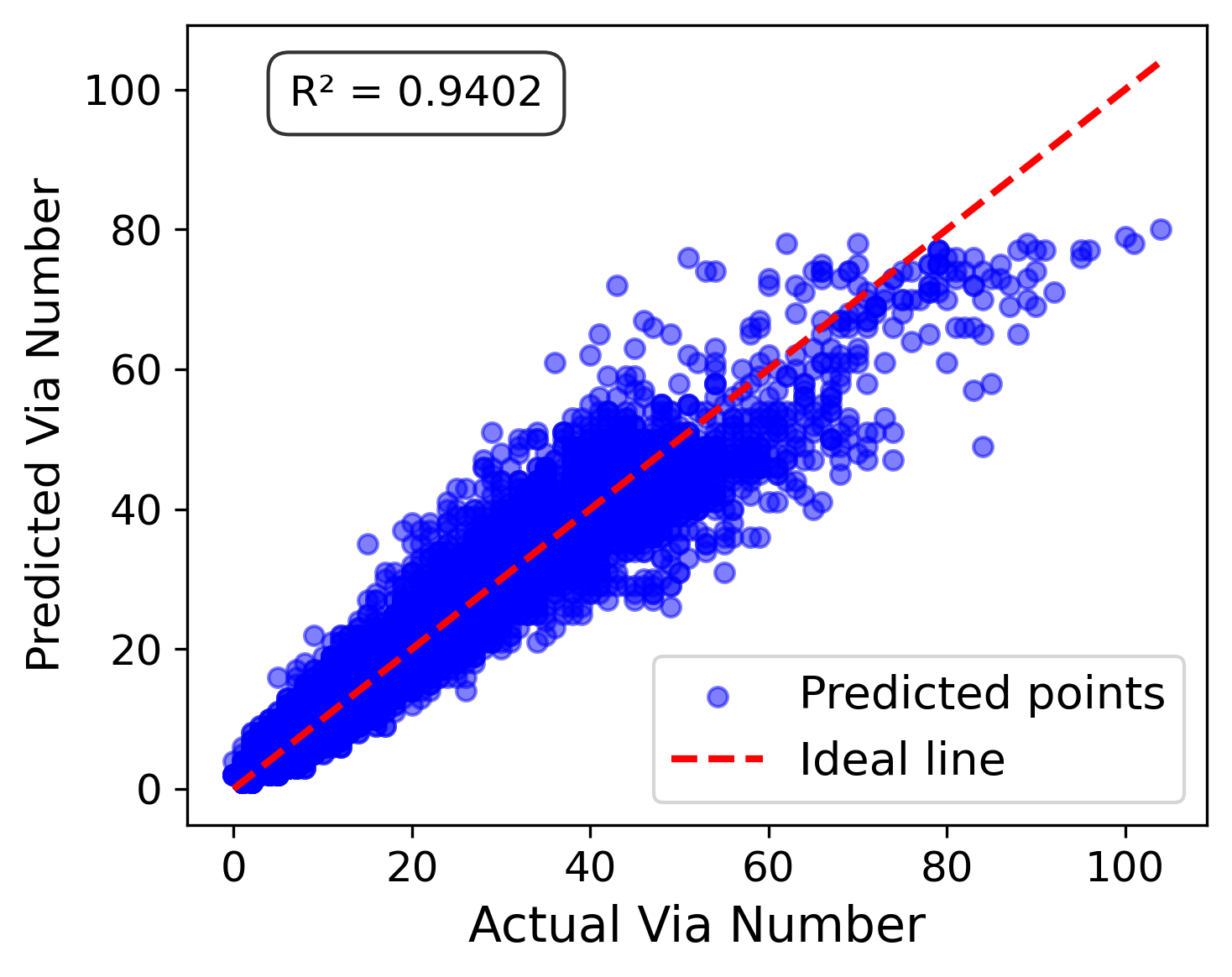}
    }
    \hfill
    \subfigure[]{
        \includegraphics[width=0.45\linewidth]{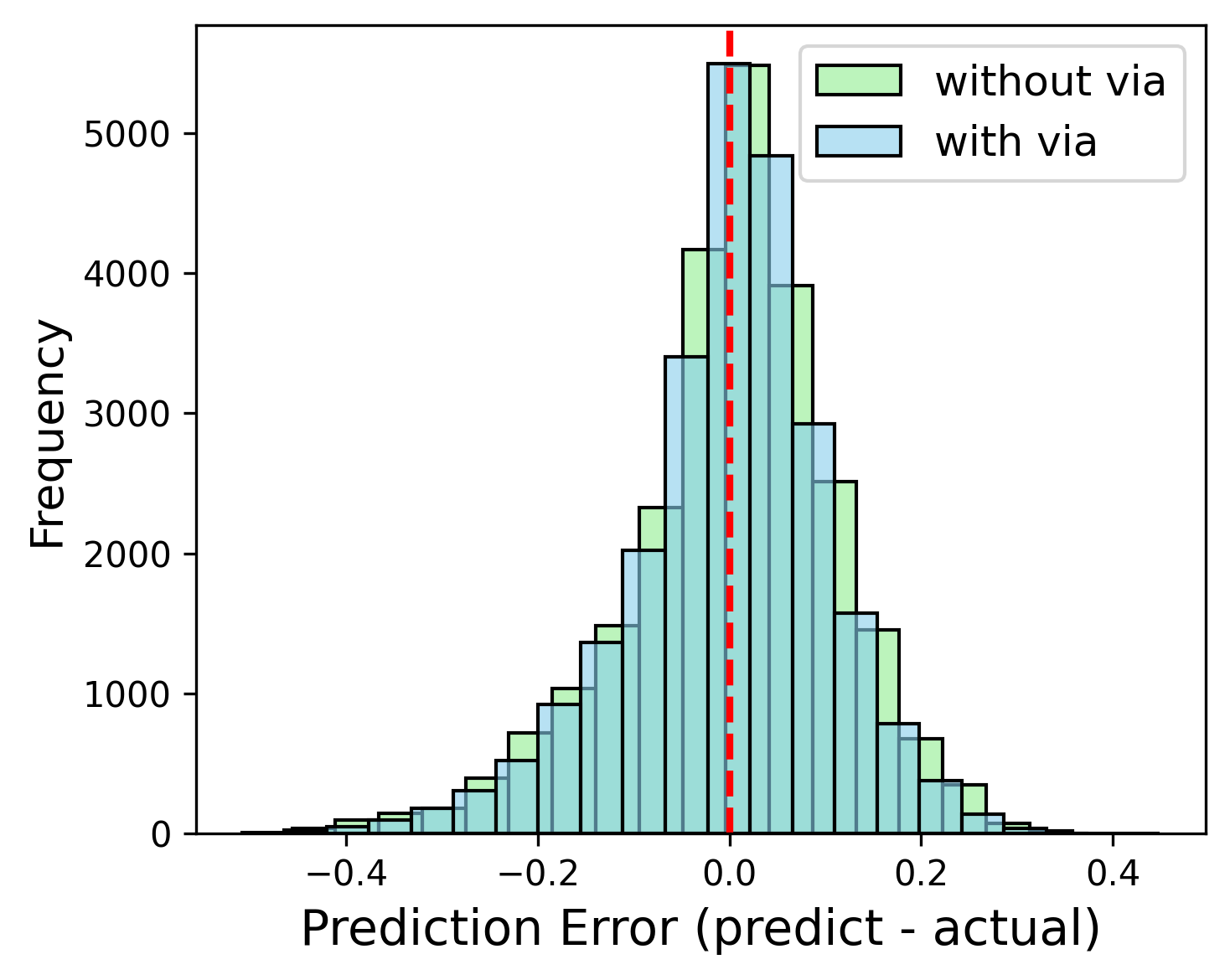}
    }
    \caption{Net level wirelength prediction results. (a) Via count prediction. (b) Error distribution with/without via features.}
    \label{fig:wirelength_predict}
\end{figure}

\subsection{Net Level Wirelength Prediction}
\label{sec:task1}
\subsubsection{Methodology}

This task estimates the wirelength ratio (RWL/RSMT) during placement. We construct the dataset from the top 30 circuit designs in \cref{tab:data}, merging their nets and randomly splitting into 80\% training (\textgreater100,000 nets) and 20\% testing (\textgreater25,000 nets). Each net sample is organized as tabular data with corresponding features and labels.

We employ TabNet\cite{arikTabNetAttentiveInterpretable2020} as the base model for its superior tabular data handling and feature selection capabilities. Our approach implements a two-stage prediction framework: (1) predicting via count using placement-stage features (aspect ratio, fanout, HPWL, RSMT, L-ness\cite{kahngWotAnalysisReal2018}), and (2) predicting wirelength ratio using both placement-stage features and the predicted via count as input. Both models are trained with mean squared error (MSE) loss.

\subsubsection{Experimental Results}
As shown in ~\cref{fig:wirelength_predict}(a), our via count prediction achieves robust   performance with $R^2 = 0.94$. ~\cref{fig:wirelength_predict}(b) compares error distributions for wirelength ratio models with and without via count features. The leftward shift indicates improved accuracy. Specifically, incorporating via prediction reduces mean relative error (MRE) by 6\%. This validates our framework's effectiveness in leveraging intermediate physical design knowledge to enhance prediction performance.

{Notably, the trained wirelength prediction model can be exported to the standard ONNX format and integrated into a C++-based detailed placement engine (e.g., iEDA) via ONNX Runtime. During optimization, the engine can then query the ONNX model in real-time, replacing low-fidelity heuristics like HPWL with a far more accurate wirelength evaluation for potential cell swaps. This workflow exemplifies AiEDA's function as a critical \textbf{bridge}, connecting AI models trained on its vectorized data directly back into the physical design loop for \textbf{online optimization}. Enabling such closed-loop, AI-driven optimization is a core objective of our future work.}

\subsection{Path Level Delay Prediction}
\label{sec:tasks:task1}
\subsubsection{Methodology}

This task predicts total path delay using parasitic R, C, and slew as features for timing analysis. We collect data from the top 30 designs in \cref{tab:data}, with each design contributing up to 3,000 samples. The first 21 designs form the training set, and the remaining 9 constitute the test set. Each path sample is structured as sequential data containing R, C, slew, and incremental delay sequences.

\begin{figure}
    \centering
    \includegraphics[width=1\linewidth]{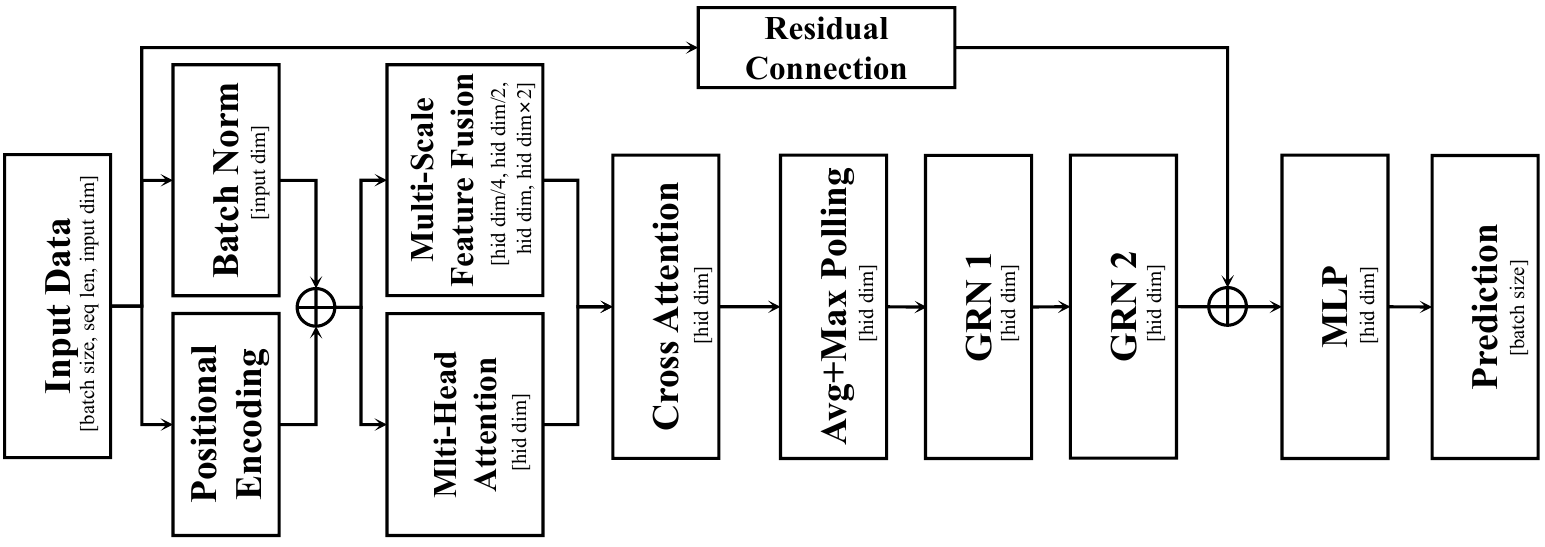}
    \caption{Transformer framework for path delay prediction.}
    \label{fig:path_delay_flow}
\end{figure}

\cref{fig:path_delay_flow} illustrates our Transformer framework. The model processes normalized input features through a hierarchical multi-scale fusion module:
\begin{equation}
    \mathbf{H}_{\text{multi}} = \sum_{d \in \left\{\frac{d}{4}, \frac{d}{2}, d, 2d\right\}} \text{Conv1D}_d(\mathbf{X}_{\text{norm}})
\end{equation}
where convolutional kernels extract features at varying temporal resolutions, subsequently fused through gated aggregation. The architecture employs dual attention pathways:
self-attention for intra-sequence modeling and cross-attention for context integration:
\begin{gather}
    \mathbf{H}_{\text{self}} = \text{LayerNorm}(\text{MHA}(\mathbf{Q}, \mathbf{K}, \mathbf{V}) + \mathbf{H}_{\text{embed}}) \\
    \mathbf{H}_{\text{cross}}^l = \text{MHA}_l(\mathbf{H}^{l-1}, \mathbf{H}_{\text{multi}})
\end{gather}
where \( l \) denotes layer depth for progressive temporal refinement. The system employs two-stage feature refinement using Gated Residual Networks (GRN) with Gated Linear Unit activation for selective information propagation. The final prediction combines refined features with adaptive residuals using Layer Normalization, preserving original signal characteristics while learning complex nonlinear mappings.

\subsubsection{Experimental Results}
\begin{figure}[t]
    \centering
    \subfigure[]{
        \includegraphics[width=0.46\linewidth]{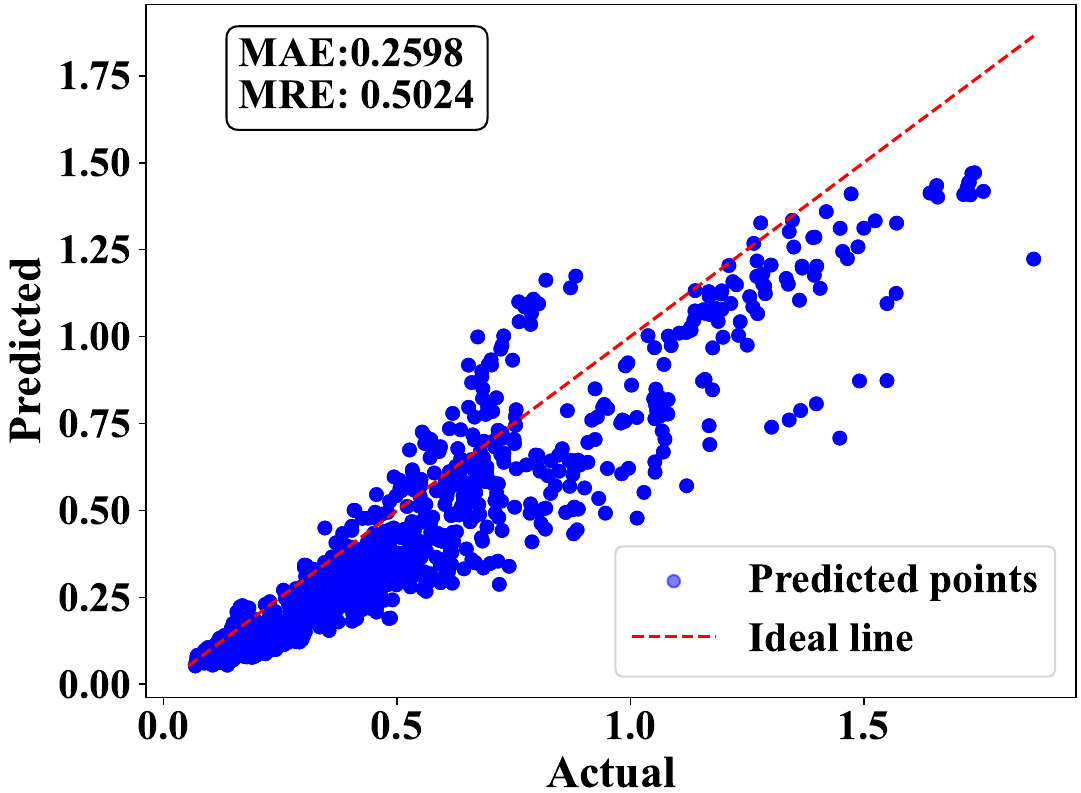}
    }
    \hfill
    \subfigure[]{
        \includegraphics[width=0.46\linewidth]{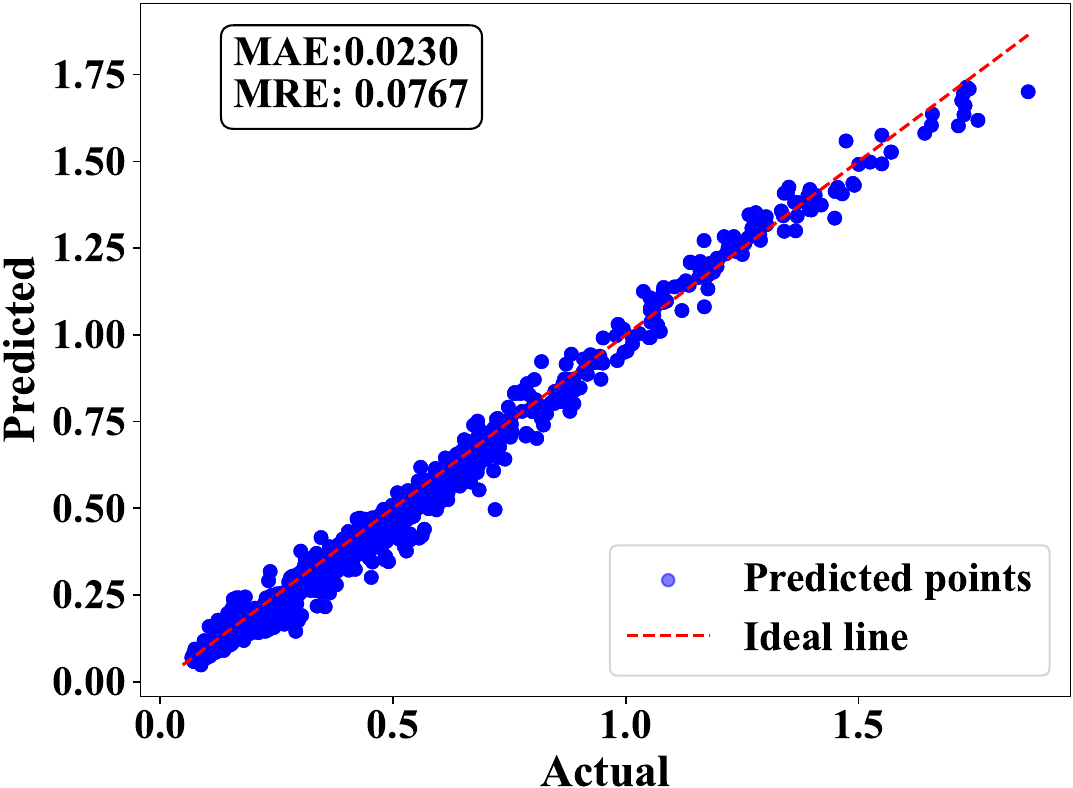}
    }
    \caption{Path level delay prediction results. (a) MHCA model accuracy (baseline). (b) MHCA+GRN model accuracy.}
    \label{fig:path_delay_predict}
\end{figure}

\cref{fig:path_delay_predict} shows the path level delay prediction results using mean absolute error (MAE) and MRE as evaluation metrics. The Multi-Head Cross Attention with GRN (MHCA+GRN) significantly outperforms the baseline MHCA model, achieving substantial improvements in both training (MAE: 0.0268 vs 0.2555, MRE: 0.1029 vs 0.5087) and testing (MAE: 0.0230 vs 0.2598, MRE: 0.0767 vs 0.5024). These results validate the effectiveness of incorporating GRNs for enhanced delay prediction accuracy.

\iftrue
\begin{algorithm}[t]
\caption{Path Delay Prediction}
\label{alg:prediction}
    \begin{algorithmic}[1]
        \STATE Input: features \( \mathbf{X} \in \mathbb{R}^{n \times d} \)
        \STATE \( \mathbf{X}_{\text{norm}} \gets \text{BatchNorm1D}(\mathbf{X}) \)
        \STATE \( \mathbf{H}_{\text{multi}} \gets \text{MultiScaleFusion}(\mathbf{X}_{\text{norm}}) \)
        \STATE \( \mathbf{H}_{\text{embed}} \gets \text{Linear}(\mathbf{X}_{\text{norm}}) + \text{PosEnc} \)
        \STATE \( \mathbf{H}_{\text{self}} \gets \text{SelfAttention}(\mathbf{H}_{\text{embed}}) \)
        \FOR{\( l = 1 \) to \( L \)}
            \STATE \( \mathbf{H}^l \gets \text{CrossAttention}(\mathbf{H}^{l-1}, \mathbf{H}_{\text{multi}}) \)
        \ENDFOR
        \STATE \( \mathbf{H}_{\text{pooled}} \gets \text{MeanPool}(\mathbf{H}^L) + \text{MaxPool}(\mathbf{H}^L) \)
        \STATE \( \hat{\mathbf{y}} \gets \text{OutputLayer}(\text{GRN}_2(\text{GRN}_1(\mathbf{H}_{\text{pooled}}))) \)
        \RETURN \( \hat{\mathbf{y}} \)
    \end{algorithmic}
\end{algorithm}
\fi

{
\subsection{Graph Level Delay Prediction}
\label{sec:tasks:task2}
\subsubsection{Methodology}

This task predicts node‐wise incremental delays on locally‐extracted critical‐path subgraphs. We extract these subgraphs from the top-30 designs listed in \cref{tab:data} by aggregating all timing paths that shared a common clock source. Vertices and edges are annotated with their corresponding physical and electrical attributes (e.g., coordinates, capacitance, resistance, slew). For training, node delays are computed, log-transformed, and then normalized on a per-design basis. The final dataset is split into training (70\%), validation (15\%), and test (15\%) sets, ensuring that all paths from a single design reside in the same split.

As illustrated in \cref{fig:graph_delay_model_1}, our approach employs a GNN-Transformer architecture. The model processes annotated netlists where nodes and edges are represented by feature vectors. First, a configurable GNN encoder (GCN, GraphSAGE, or GIN) captures local topological and electrical context, producing initial node embeddings. Second, these embeddings are augmented with sinusoidal graph-Laplacian positional encodings and fed into a Transformer encoder. This allows the model to capture long-range dependencies along timing paths via its multi-head self-attention mechanism. Finally, a lightweight MLP maps the final node embeddings to the predicted delay values.

\begin{figure}[t]
    \centering
    \includegraphics[width=1\linewidth]{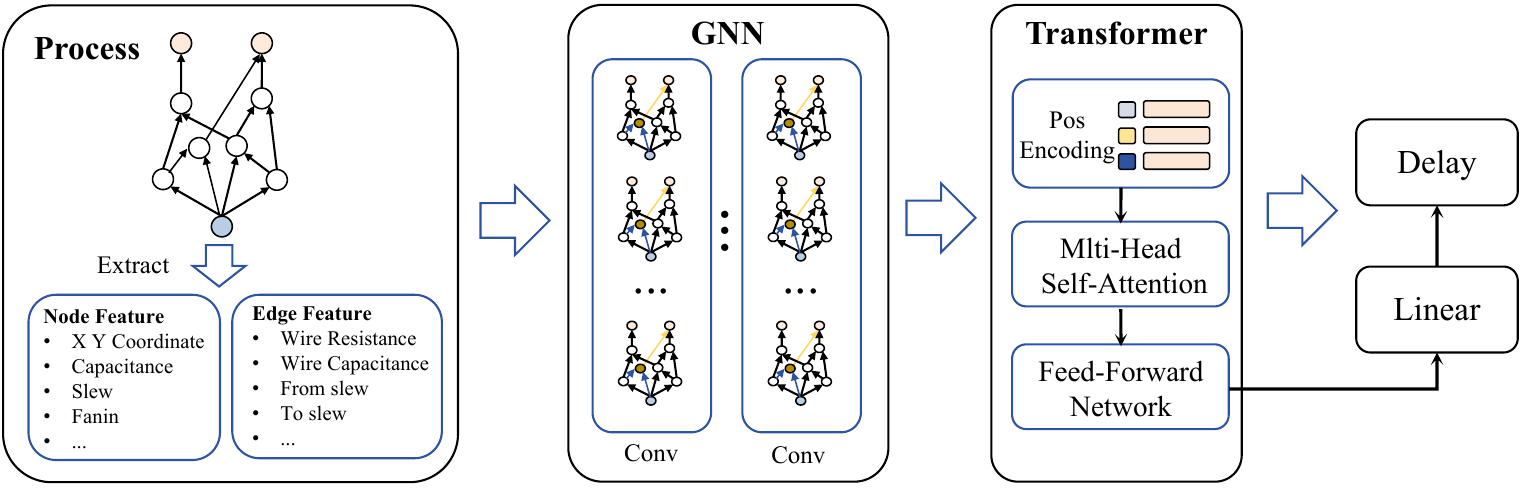}
    \caption{{Graph level delay prediction framework.}}
    \label{fig:graph_delay_model_1}
\end{figure}

\subsubsection{Experimental Results}


\begin{figure}[t]
    \centering
    \includegraphics[width=1\linewidth]{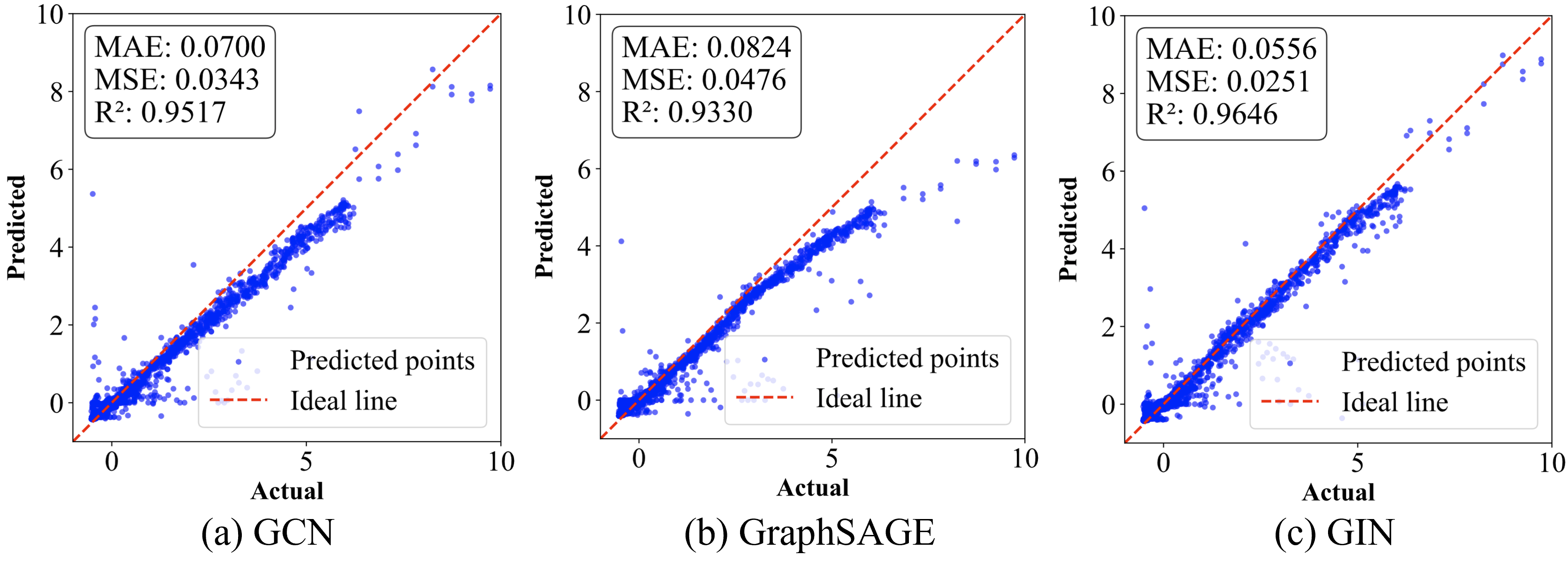}
    \caption{{Scatter plot comparison of the prediction results of three GNNs combined with Transformer on the test data.}}
    \label{fig:Comparison_graph_delay_1}
\end{figure}

\begin{table}[!t]
\centering
\caption{Comparison of test-set performance among GCN, GraphSAGE and GIN. 
         Best results are highlighted in \textbf{bold}.}
\label{tab:graph_delay_results}
\setlength{\tabcolsep}{14.5pt}
\begin{tabular}{lccc}
\toprule
Model & MSE $\downarrow$ & MAE $\downarrow$ & R\textsuperscript{2} $\uparrow$ \\
\midrule
GCN   & 0.034274 & 0.070015 & 0.951728 \\
GraphSAGE  & 0.047566 & 0.082359 & 0.933007 \\
GIN   & \textbf{0.025113} & \textbf{0.055604} & \textbf{0.964630} \\
\bottomrule
\end{tabular}
\end{table}

The performance and the prediction scatter plots of the three GNN encoders on the test set \idel{is summarized in Table~\ref{tab:graph_delay_results_1} and \cref{tab:graph_delay_results}, with corresponding prediction scatter plots}are shown in \cref{fig:Comparison_graph_delay_1}. All models achieve strong results (R² $>$ 0.93), demonstrating the effectiveness of GNN-based architectures for timing prediction.
Among them, GIN delivers state-of-the-art performance, achieving an MSE of 2.51\% and an R² of 0.9646. This represents a substantial improvement over the next-best model, GCN, with 26.7\% lower MSE and 20.6\% lower MAE. This advantage stems from GIN's powerful injective aggregation function, which allows it to distinguish subtle local sub-structures that influence delay. Conversely, the simple mean pooling of GraphSAGE proves less effective, resulting in the highest error and underscoring its inadequacy for tasks requiring sign-off-level precision.

}

\vspace{-0.2cm}
\subsection{Patch Level Congestion Prediction}
\label{sec:tasks:task3}
\subsubsection{Methodology}

{To predict the early global routing (EGR) congestion map, we use a set of features that are all obtainable from the placement stage, including cell density, pin density, net density, and RUDY\cite{spindlerFastAccurateRouting2007e}.} {We construct our dataset from the top 30 designs listed in \cref{tab:data}. As the number of patches per design varies significantly, a simple random split is suboptimal. To ensure our training, validation, and test sets are representative of this diversity, we employ a stratified sampling strategy. We group the designs into three strata based on patch count (large, medium, and small) and then randomly sample from each stratum to create a 19-design training set, a 3-design validation set, and an 8-design test set.}

\begin{figure}[t]
    \centering
    \includegraphics[width=1.03\linewidth]{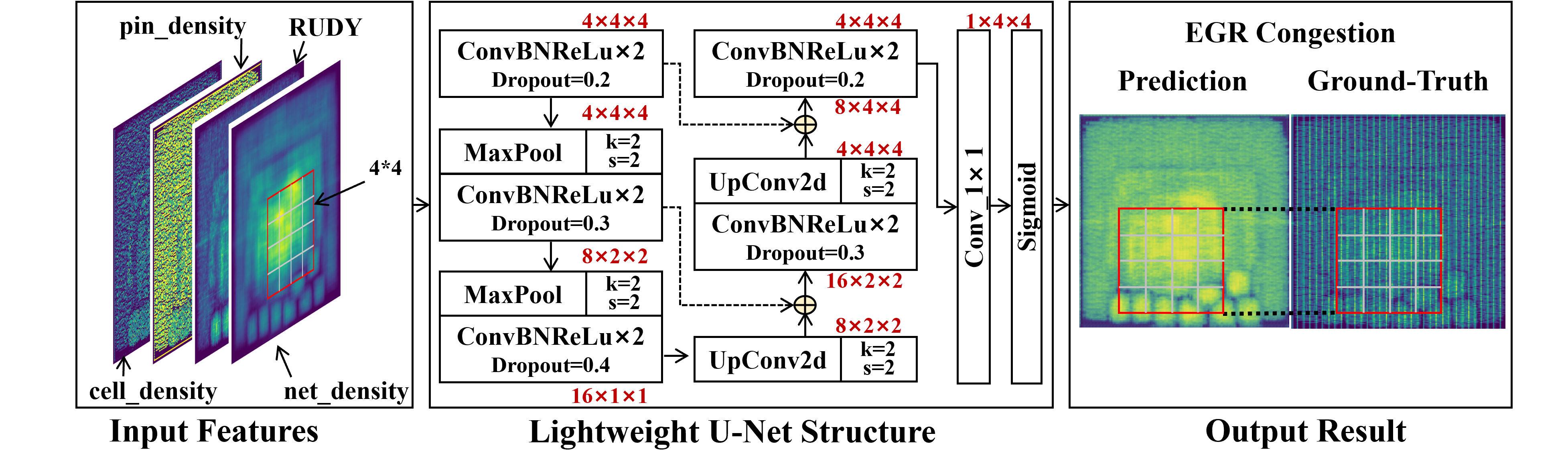}
    \caption{U-Net architecture with multi-feature input and sliding window processing for patch level congestion prediction.}
    \label{fig:U-net_Structure}
\end{figure}

Fig.~\ref{fig:U-net_Structure} illustrates our data processing and baseline model architecture. We implement a sliding window approach with 4×4 patches and stride of 3 to extract input features, effectively capturing local spatial information and addressing inconsistent input dimensions across designs. After standard normalization, we design a lightweight U-Net model with end-to-end training using MSE loss for pixel-wise congestion regression.

\begin{figure}[t]
    \centering
    \subfigure[]{
        \includegraphics[width=0.46\linewidth]{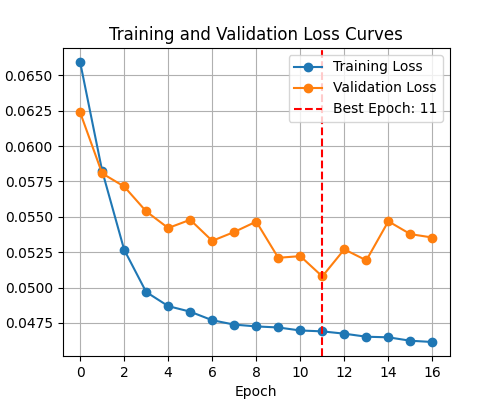}
    }
    \hfill
    \subfigure[]{
        \includegraphics[width=0.46\linewidth]{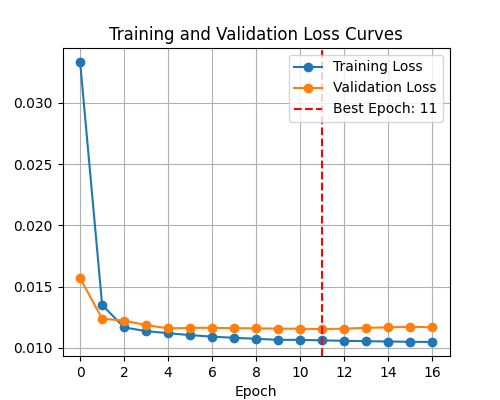}
    }
    \caption{Loss curves for different optimization strategies. (a) Robust normalization. (b) Enhanced model.}    
    \label{fig:path_congestion_predict}
\end{figure}

\begin{figure}[!t]
    \centering
    \includegraphics[width=0.8\linewidth]{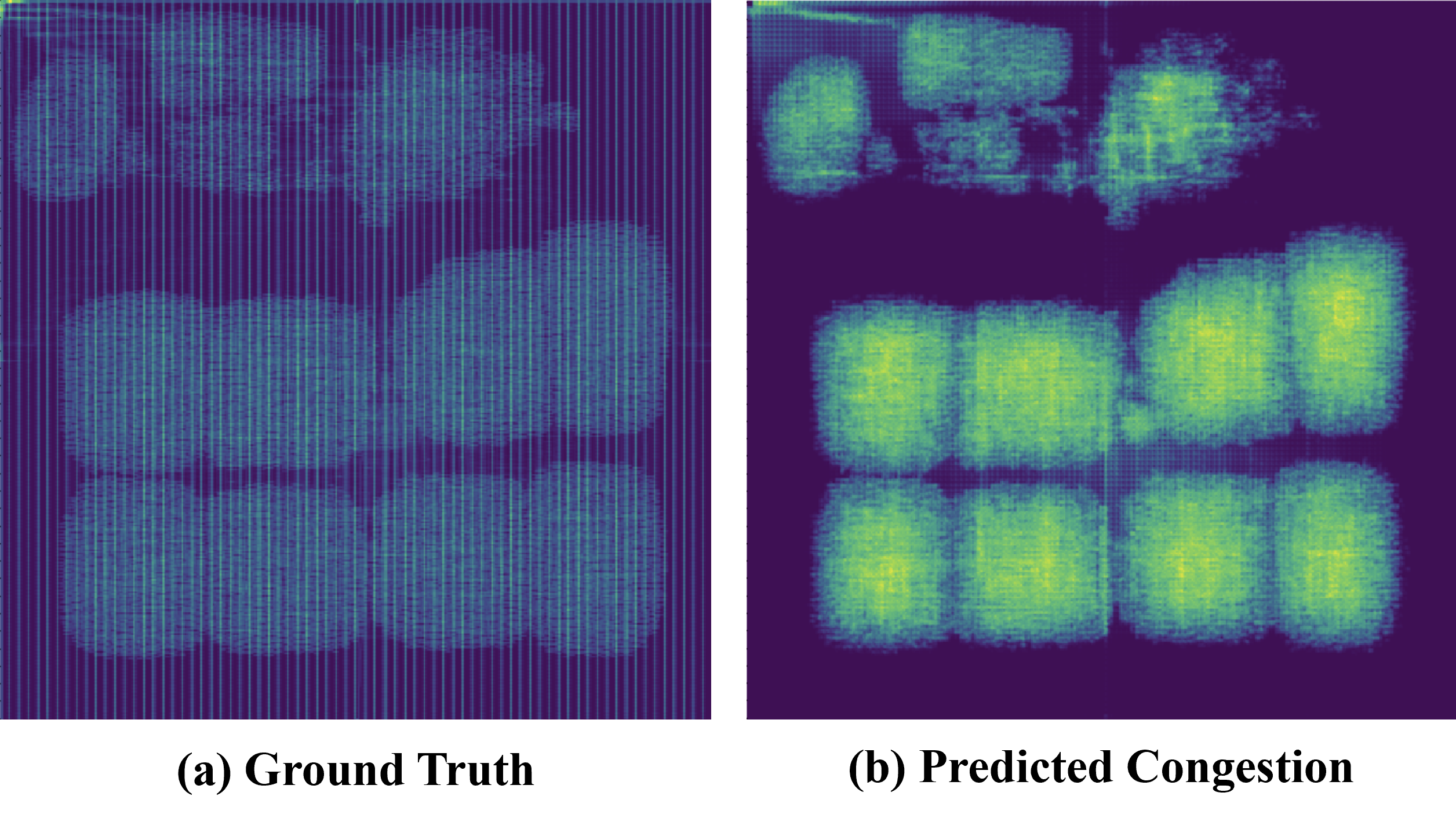}
    \caption{Comparison of the ground truth EGR congestion map (left) and the model's prediction (right) for the large-scale \texttt{eth\_top} design. The model achieves an NRMSE of 0.12 on this task. The low NRMSE achieved on this large and complex design underscores the model's effectiveness and generalizability, which are now validated on a more credible and representative test set.}
    \label{fig:eth_top_congestion_comp}
\end{figure}

\subsubsection{Experimental Results}
{The model achieves an average normalized root mean squared error (NRMSE) of 0.18 on the test set. While the model tends to overestimate congestion, it accurately preserves relative spatial patterns, which is sufficient for congestion-aware optimization. We further conduct two comparative experiments to evaluate optimization strategies.  Fig.~\ref{fig:path_congestion_predict}(a) shows results using robust normalization (RobustScaler), yielding a suboptimal average NRMSE of 0.23. Fig.~\ref{fig:path_congestion_predict}(b) demonstrates improved results with an average NRMSE of 0.18, achieved through increased model capacity (expanding channels from 4→8→16 to 16→32→64) and attention mechanisms. First, using robust normalization (RobustScaler) resulted a suboptimal average NRMSE of 0.23. In contrast, enhancing the model capacity (expanding channels from 4→8→16 to 16→32→64) successfully reduced the average NRMSE to 0.17.} {We observe that the model performs notably better on medium and large-scale designs, achieving NRMSE values as low as 0.12, as shown in \cref{fig:eth_top_congestion_comp}. This improved performance is likely attributable to the richer and more diverse congestion patterns present in larger layouts, which provide more comprehensive learning opportunities for the model.}

\subsection{Net and Map Level Routing {Mask} Generation}
\label{sec:tasks:task4}
\subsubsection{Methodology}
This task generates a two-pin net routing map using net-level data (source and target points) with {4} spatial patch features (cell density, pin density, net density, RUDY) in a multi-modal representation. {These features are all obtainable from the placement stage.} \idel{We employ 20 designs from the \texttt{apb4} series and \texttt{s} series as our dataset, partitioned at the design level: 16 designs for training (5,179 samples), 2 for validation (1,469 samples), and 2 for testing.}{We employ 20 designs from the \texttt{apb4} series and \texttt{s} series as our dataset. Following the stratified sampling methodology detailed in \cref{sec:tasks:task3}, the dataset is partitioned into a 12-design training set, a 3-design validation set, and a 5-design test set.}

Net-level source and target points undergo one-hot encoding. Patch features are processed through intra-design normalization and spatial relative feature computation, yielding a {8}-dimensional composite patch feature vector. All patch regions are resized to 16×16 grids for standardized processing.

We adopt a U-Net architecture as our base model. The encoder and decoder paths are represented as:
\begin{equation}
\{\mathbf{E}_1, \mathbf{E}_2, \mathbf{B}\} = \text{Encoder}(\mathbf{X}; {10} \rightarrow 16 \rightarrow 32 \rightarrow 64)
\end{equation}
\begin{equation}
\mathbf{Y} = \text{Decoder}(\mathbf{B}, \{\mathbf{E}_1, \mathbf{E}_2\}; 64 \rightarrow 32 \rightarrow 16 \rightarrow 1)
\end{equation}
The encoder progressively downsamples through double convolution blocks and max pooling to extract multi-scale features, while the decoder upsamples via transposed convolution and combines skip connections to recover spatial details. The input $\mathbf{X} \in \mathbb{R}^{{10} \times 16 \times 16}$ contains {8}-dimensional composite features and 2-dimensional source-target encodings, producing output $\mathbf{Y} \in \mathbb{R}^{1 \times 16 \times 16}$ representing the path probability map. The model is trained using binary cross-entropy loss:
\begin{equation}
\mathcal{L} = -\frac{1}{N} \sum_{i=1}^{N} [y_i \log(\sigma(\hat{y}_i)) + (1-y_i) \log(1-\sigma(\hat{y}_i))]
\end{equation}
where $N$ is the total number of patches, $y_i$ is the ground truth, $\hat{y}_i$ is the predicted logit value, and $\sigma$ is the sigmoid function.

\subsubsection{Experimental Results}
\begin{figure}[t]
\centering
\includegraphics[width=1\linewidth]{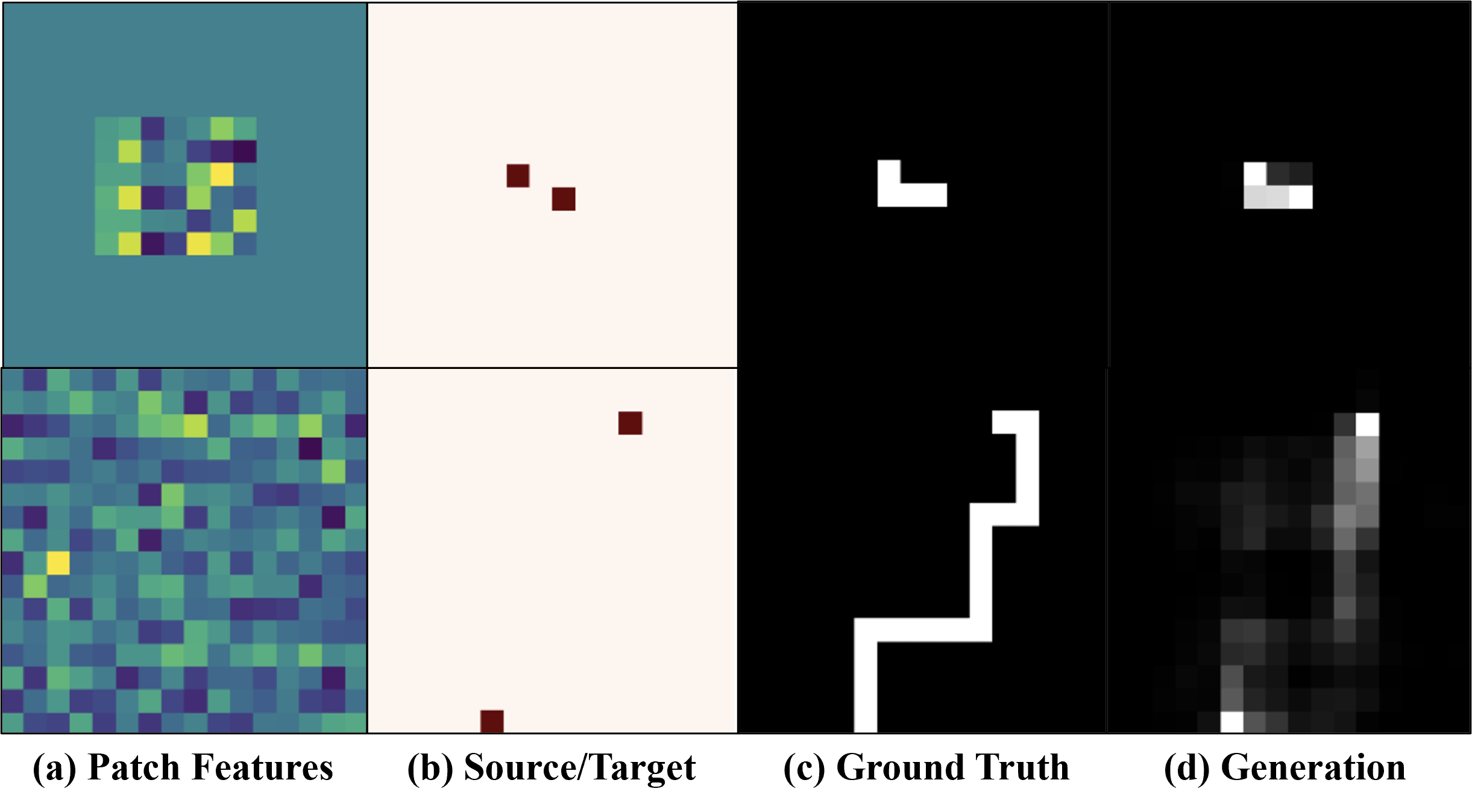}
\caption{{Qualitative results of net map generation for short and long paths.}}
\label{fig:routing_mask_flow}
\end{figure}

{The model generates a routing probability map for each patch. For evaluation, these probabilities are binarized using a threshold (default = 0.4), where 1 signifies a generated path. On the test set, the model achieves an 81\% F1-score and a 67\% Intersection over Union (IoU), indicating a reasonable spatial overlap with the ground truth. Fig.~\ref{fig:routing_mask_flow} demonstrates routing generation for nets with both short and long path lengths. As illustrated, for short-distance nets, the generated path closely matches the ground truth. However, for long-distance nets, while the model correctly captures the overall trajectory, the lower probabilities in the intermediate segments can result in discontinuous paths upon binarization. In a practical application scenario, the raw (i.e., non-binarized) probability maps from multiple nets can be aggregated to estimate overall routing density, which provides predictive insights at earlier design stages; for instance, by serving as a fast congestion predictor to guide routability-driven placement or acting as a look-ahead engine in early-stage global routing to anticipate resource contention. The result of routing mask generation conbined with net resource map can be used to guide the net routing, as shown in \cref{fig:ai_assisted_gr_flow}.

\begin{figure}[!t]
    \centering
    \includegraphics[width=1\linewidth]{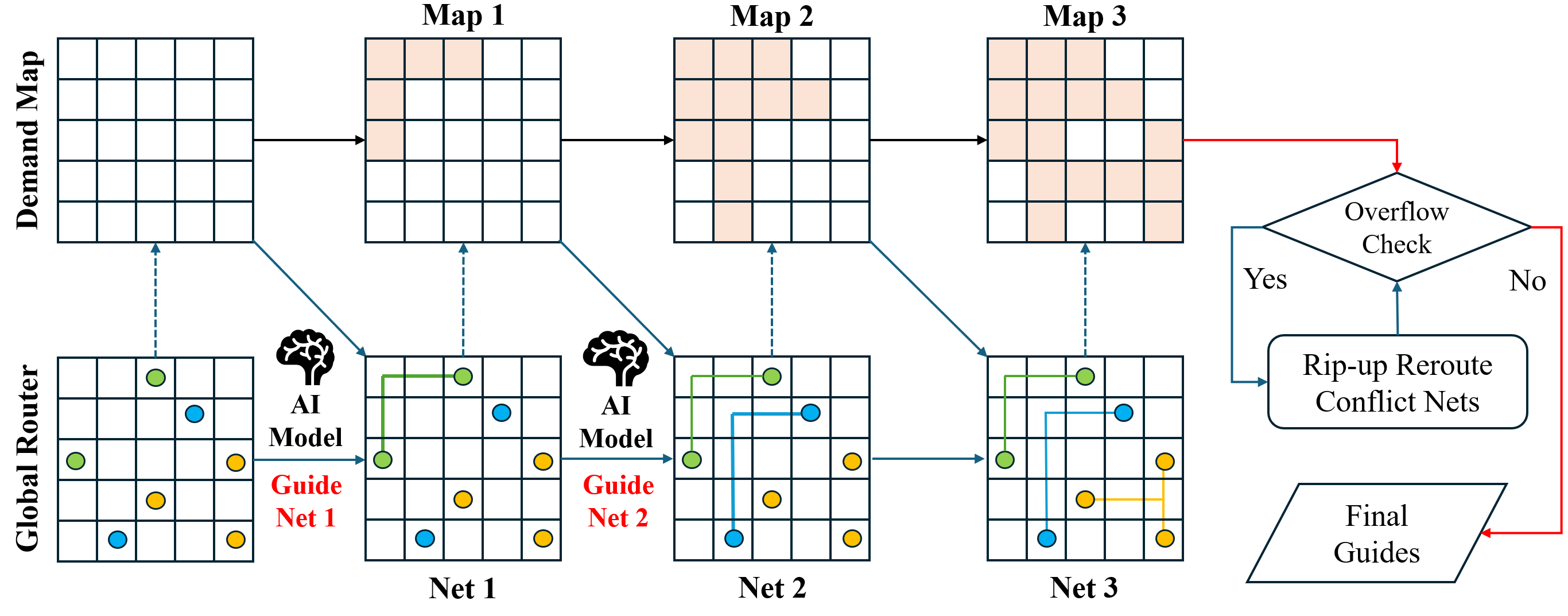}
    \caption{Illustration of the proposed AI-assisted routing flow. By leveraging an AI model for look-ahead resource prediction, the router anticipates that the default paths for Net 1 and Net 2 would create a future bottleneck. Guided by this foresight, the router proactively adjusts their paths (Guide Net 1 and Guide Net 2). This preemptive rerouting successfully preserves critical resources, prevents the overflow, and eliminates the need for RRR iterations, thereby accelerating overall design convergence.}
    \label{fig:ai_assisted_gr_flow}
\end{figure}

\subsection{Design Level Parameter Optimization}
\label{sec:tasks:task5}
\subsubsection{Methodology}

This task optimizes tool parameters to achieve superior performance across design stages. We select the iEDA placement engine as the optimization target. The selected parameters of iEDA placement are listed in \cref{tab:parameter_priors}.
We employ the multi-objective tree-structured Parzen estimator (MOTPE) algorithm to optimize key placement metrics, including HPWL, WNS, and TNS. These metrics are extracted from design-level evaluation files (\texttt{eval.json}). In each iteration, MOTPE predicts parameter values that maximize expected improvement (EI) based on historical parameter-metric pairs. The predicted parameters are then applied to the placement tool to generate evaluation metrics. Subsequently, the new parameters and corresponding metrics are added to the historical dataset. We set the iteration number to 100. Upon completion, we collect metrics on the Pareto frontier as experimental results.

\begin{table}[t]
\centering
\small
\caption{Parameters and their prior distributions.}
\label{tab:parameter_priors}
\setlength{\tabcolsep}{2pt}
\begin{tabular}{p{3.4cm}|p{2.6cm}|p{1.1cm}|p{1.1cm}}
\toprule
\textbf{Parameter} & \textbf{Prior Distribution} & \textbf{Default Value}  & \textbf{Tuned Value}\\
\midrule
init\_wirelength\_coef & $U(0.1, 0.5)$ & $0.250$  & $0.187$  \\
min\_wirelength\_force\_bar & $U(-500.0, -50.0)$ & $-300$  & $-341.7$  \\
target\_density & $U(0.8, 1.0)$ & $0.80$ & $0.89$ \\
bin\_cnt & \{16, 32, 64, 128, 256, 512, 1024\} & $64$ & $128$\\
max\_backtrack & $U(5, 50)$ & $10.0$ & $33.7$\\
init\_density\_penalty & $U(0.0, 0.001)$ & $0.00008$ & $0.00051$\\
target\_overflow & $U(0.0, 0.2)$ & $0.1$ & $0.0446$\\
initial\_prev\_coordi\_coef & $U(50.0, 1000.0)$ & $100$ & $241.3$\\
min\_precondition & $U(1.0, 10.0)$ & $1.0$ & $2.4$\\
min\_phi\_coef & $U(0.75, 1.25)$ & $0.95$ & $0.87$\\
max\_phi\_coef & $U(0.75, 1.25)$ & $1.05$ & $1.03$\\
\bottomrule
\multicolumn{2}{l}{$U$ denotes uniform distribution.}
\end{tabular}
\end{table}

\begin{table}[t]
\centering
\caption{Comparison of default and optimized parameters across different designs.}
\label{tab:placement_dse_comparison}
\setlength{\tabcolsep}{4pt}
\begin{tabular}{l|rr|rr|rr}
\toprule
\multirow{2}{*}{Design} & \multicolumn{2}{c|}{HPWL $\downarrow$} & \multicolumn{2}{c|}{WNS $\uparrow$} & \multicolumn{2}{c}{TNS $\uparrow$} \\
\cmidrule(lr){2-3} \cmidrule(lr){4-5} \cmidrule(lr){6-7}
 & \multicolumn{1}{c}{Default} & \multicolumn{1}{c|}{Tuning} & \multicolumn{1}{c}{Default} & \multicolumn{1}{c|}{Tuning} & \multicolumn{1}{c}{Default} & \multicolumn{1}{c}{Tuning} \\
\midrule
\texttt{s713} & 1.24M& \textbf{1.17M} & -0.84 & \textbf{0.33} & -4.78 & \textbf{0.00} \\
\texttt{s1238} &\textbf{ 2.69M} & 2.77M& -0.05 & \textbf{0.29} & -0.05 & \textbf{0.00} \\
\texttt{s1488} & \textbf{3.51M} & 3.67M & -0.20 & \textbf{0.16} & -0.87 & \textbf{0.00} \\
\texttt{s9234} & 6.74M & \textbf{5.30M} & -0.30 & \textbf{-0.09} & -2.33 & \textbf{-1.16} \\
\texttt{s13207} & 5.94M & \textbf{5.81M} & 0.44 & \textbf{0.46} & 0.00 & \textbf{0.00} \\
\texttt{apb4\_ps2} & 4.76M & \textbf{4.00M}& 0.09 & \textbf{0.90} & 0.00 & \textbf{0.00} \\
\texttt{apb4\_timer} & 9.81M & \textbf{6.99M}& 0.13 & \textbf{0.59} & 0.00 & \textbf{0.00} \\
\texttt{apb4\_i2c} & 10.11M & \textbf{6.60M} & 0.12 & \textbf{0.65} & 0.00 & \textbf{0.00} \\
\texttt{apb4\_pwm} & 13.77M & \textbf{8.92M} & 0.21 & \textbf{0.63} & 0.00 & \textbf{0.00} \\
\texttt{apb4\_wdg} & 14.74M & \textbf{9.69M} & 0.10 & \textbf{0.51} & 0.00 & \textbf{0.00} \\
\midrule
\textbf{Impr. Rate} & \textbf{—} & \textbf{8/10} & \textbf{—} & \textbf{10/10} & \textbf{—} & \textbf{10/10} \\
\bottomrule
\end{tabular}
\end{table}

\begin{table}[t]
\centering
\caption{Single-objective (HPWL) optimization on larger designs.}
\label{tab:large_scale_dse_1}
\setlength{\tabcolsep}{6.7pt}
\begin{tabular}{l|r|rr|c}
\toprule
\textbf{Design} & \textbf{\# Cells} & \multicolumn{1}{c}{\textbf{Default HPWL}} & \multicolumn{1}{c|}{\textbf{Tuned HPWL}} & \textbf{Impr.} \\
\midrule
\texttt{jpeg} & 27.7k & 1118.1M & \textbf{312.6M} & 72\% \\
\texttt{eth\_top} & 42.3k & 1410.3M & \textbf{691.4M} & 51\% \\
\texttt{yadan} & 63.5k & 1469.2M & \textbf{1155.7M} & 21\% \\
\texttt{SHMS} & 268.7k & 5669.5M & \textbf{4661.5M} & 18\% \\
\texttt{nvdla} & 289.3k & 26677.6M & \textbf{14253.1M} & 47\% \\
\bottomrule
\end{tabular}
\end{table}

\begin{figure*}[t]
    \centering
    \includegraphics[width=\linewidth]{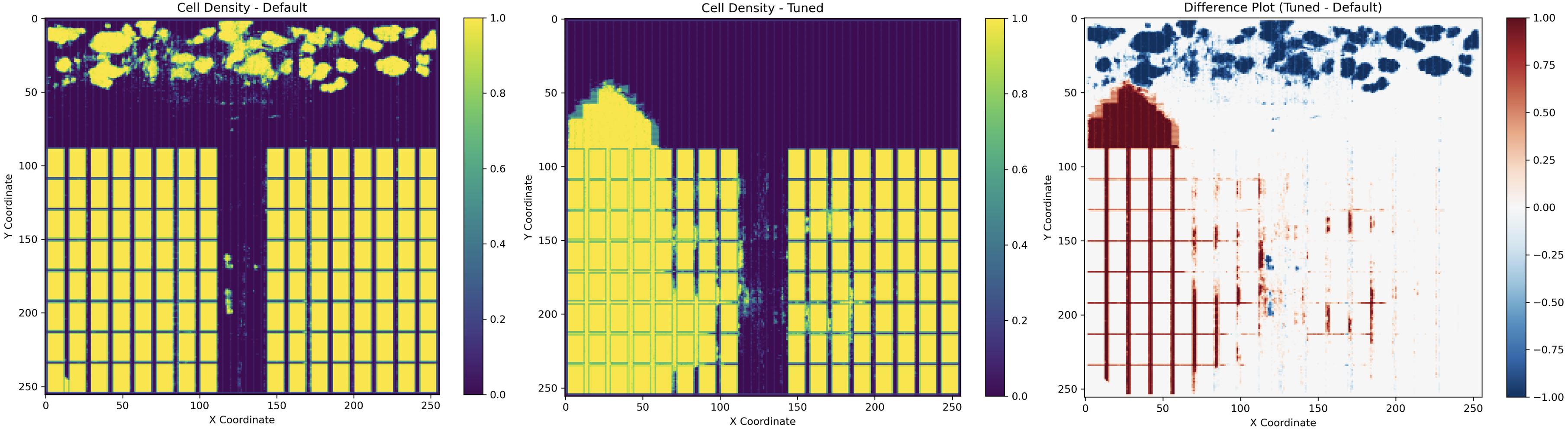}
    \caption{Cell density heatmaps for the ``nvdla" design before (left) and after (middle) single-objective HPWL optimization. The difference plot (right) visually confirms this dramatic shift in cell placement.}
    \label{fig:cell_density_heatmap_comparison_1}
\end{figure*}

\subsubsection{Experimental Results}
\cref{tab:placement_dse_comparison} compares default settings with optimized parameters across multiple designs. Our approach achieves improvements in HPWL (8/10 designs), WNS (10/10 designs), and TNS (10/10 designs), validating the effectiveness of our parameter optimization methodology. 
{To demonstrate the scalability of our methodology, we simplified the problem to a single-objective optimization (HPWL) and ran 100 DSE iterations for several large-scale designs, as shown in \cref{tab:large_scale_dse_1}. These new results demonstrate that our DSE methodology is highly effective on large designs, achieving significant wirelength improvements (up to 72\%).  \cref{fig:cell_density_heatmap_comparison_1} shows the cell density for ``nvdla" before and after tuning.} 
{We note that the scalability of this multi-objective DSE is currently limited by the runtime of the underlying academic placer, which is why we focused on these smaller benchmarks for the comprehensive multi-objective evaluation. A key direction for future work is to integrate this DSE framework with a high-performance placer, such as DREAMPlace, to enable large-scale multi-objective optimization.}

\iftrue
\subsection{Design Level Metric Analysis and Tool Comparison}
\label{sec:tasks:task6}
\subsubsection{Methodology}
The design metric analysis task analyzes performance disparities between two flow engines. We conduct a comparative analysis between the placement engines of iEDA and Innovus. The workflow generates different placement results for identical chip designs using these two placement engines, followed by completing the physical design process using iEDA. 

Our evaluation methodology employs a two-level assessment approach: intermediate metrics are evaluated using iEDA's evaluator on placement results, while PPA metrics are measured using Innovus on final routing results. We select the first 30 designs from the dataset presented in \cref{tab:data} and subject them to our workflow and performance evaluation protocol. Representative results are illustrated in \cref{tab:task6_tool_metrics}.

\begin{table}[t]
\centering
\caption{Comparison of metrics between iEDA and Innovus placement engines.}
\label{tab:task6_tool_metrics}
\setlength{\tabcolsep}{1.1pt}
\begin{tabular}{llcccccc}
\toprule
\multirow{2}{*}{Designs} & \multirow{2}{*}{Engines} & \multicolumn{2}{c}{Intermediate Metrics} & \multicolumn{4}{c}{PPA Metrics} \\
\cmidrule(lr){3-4} \cmidrule(lr){5-8} 
 & & RSMT & Overflow & Wirelength & \#DRCs & WNS & Power \\ 
 \midrule
 \multirow{2}{*}{\texttt{s15850}} & iEDA & 23.12M & 60 & 42.48M & 9,523 & -2.54 & 1.63 \\ \cline{2-8} 
 & Innovus & 25.30M & 44 & 38.32M & 5,750 & -0.61 & 1.64 \\ 
 \midrule
 \multirow{2}{*}{\texttt{picorv32}} & iEDA & 172M & 108 & 308M & 47,411 & -1.34 & 3.16 \\ \cline{2-8} 
 & Innovus & 176M & 44 & 236M & 25,893 & -0.14 & 3.33 \\ 
 \midrule
 \multirow{2}{*}{\texttt{jpeg}} & iEDA & 422M & 0 & 1281M & 373,696 & 3.66 & 9.13 \\ \cline{2-8} 
 & Innovus & 434M & 0 & 1228M & 111,956 & 4.61 & 10.35 \\ 
 \midrule
 \multirow{2}{*}{\texttt{apb4\_uart}} & iEDA & 77.66M & 10 & 165.30M & 30,271 & 0.10 & 3.84 \\ \cline{2-8} 
 & Innovus & 84.53M & 6 & 139.10M & 18,354 & 0.22 & 3.85 \\ 
 
\midrule
 \multirow{2}{*}{\texttt{s9234}} & iEDA & 5.76M & 8 & 7.36M & 2,528 & -0.54 & 0.31 \\ \cline{2-8} 
 & Innovus & 6.48M & 0 & 8.17M & 1,629 & 0.19 & 0.54 \\ 
\midrule
 \multirow{2}{*}{\texttt{s5378}} & iEDA & 9.99M & 16 & 12.83M & 3,729 & -0.52 & 0.41 \\ \cline{2-8} 
 & Innovus & 11.50M & 6 & 14.46M & 2,287 & 0.06 & 0.45 \\ 
\bottomrule
\end{tabular}
\end{table}

\subsubsection{Experimental Results}
\paragraph{Intermediate vs. Final Metrics Discrepancy}
Designs such as \texttt{s15850}, \texttt{picorv32}, \texttt{jpeg}, and \texttt{apb4\_uart} demonstrate superior intermediate metrics (RSMT) when processed through iEDA; however, their final PPA metrics (wirelength) are inferior compared to results obtained using Innovus. This counterintuitive finding underscores a critical consideration in tool performance analysis: excessive focus on intermediate metrics may lead to misleading conclusions, whereas PPA metrics provide more reliable indicators of actual performance.
\paragraph{Multi-dimensional Performance Trade-offs}
While designs such as \texttt{s9234} and \texttt{s5378} exhibit better wirelength metrics (both intermediate and PPA) when processed through iEDA, they demonstrate inferior performance on other critical metrics including DRC violations and WNS. This observation highlights the necessity of considering multiple metrics simultaneously when evaluating results, rather than relying on any single performance indicator.
\fi

Beyond the downstream tasks previously discussed, our AiEDA framework provides broader foundational support for AI-driven EDA applications. It generates essential vectorized data, offers a versatile Python API, and manages the complete neural network training lifecycle. These capabilities enable a broad spectrum of AI-enhanced EDA applications, including DRC prediction~\cite{li25aidrc}, Steiner tree generation~\cite{liu24neuralsteiner}, 3D capacitance extraction~\cite{cai24pct-cap}, timing prediction~\cite{liu25aitpo}, timing optimization~\cite{wu24aito}, IR drop calculation~\cite{liu24simultaneous}, parameter optimization~\cite{lai25ipo}, and technology mapping~\cite{liu23aimap}. {Furthermore, its multi-engine capabilities extend its utility beyond the AI community to the broader EDA community for tasks such as tool metrics benchmarking.}

\section{Conclusion}
\label{sec:conclusion}
In this work, we introduce our open-source AI-aided design (AAD) library (AiEDA) and release the iDATA dataset. 
{
AiEDA integrates multiple design-to-vector techniques, which converts chip design data into structured representations compatible with neural network model input/output pipelines. AiEDA provides unified workflows from design execution to AI integration, while iDATA contains 600GB of multi-level structured data from 50 real designs.}
Through five representative downstream tasks, we demonstrated the effectiveness of our approach across design-level, net-level, {graph/path-level}, and patch-level representations. This work provides the first unified library that addresses complete data pipeline challenges in AAD, enabling researchers worldwide to advance AI-aided design automation more effectively. {To foster collaboration, its open-source and modular architecture is designed to be highly extensible, welcoming community contributions to both the library's functionalities and the growing iDATA ecosystem.}
Future work will focus on expanding vectorization capabilities (e.g., layout-to-vector), improving dataset generation efficiency, integrating additional downstream tasks, and establishing tighter EDA flow coupling for optimization guidance. We believe this work will accelerate AI development and adoption in the EDA domain.



\bibliographystyle{IEEEtran}
\bibliography{ieda,ref}

\end{document}

%% file: ieda-config.tex
\usepackage{CJKutf8}
\usepackage{color}
\usepackage{textcomp}
\usepackage[normalem]{ulem}

\usepackage{xcolor}
    \definecolor{red}{RGB}{255,0,0}
    \definecolor{green}{RGB}{0,255,0}
    \definecolor{blue}{RGB}{0,0,255}
    \definecolor{darkgreen}{RGB}{0, 100, 0}  

\IEEEoverridecommandlockouts

\usepackage{amsmath,amssymb,amsfonts}
\usepackage{bm}

\usepackage{algorithmic}
\usepackage{algorithm}
\floatname{algorithm}{Algorithm}

\usepackage{listings}
    \lstdefinestyle{pythonstyle}{
        language=Python,
        basicstyle=\ttfamily\footnotesize,
        keywordstyle=\color{blue}\bfseries,
        commentstyle=\color{green!60!black},
        stringstyle=\color{red},
        frame=single,
        breaklines=true,
        breakatwhitespace=true,
        tabsize=4,
        showstringspaces=false,
        captionpos=b,
        framexleftmargin=-3pt, 
        framexrightmargin=-5pt
    }

\usepackage{graphicx}

\usepackage{subfigure}
\usepackage{multirow}
\usepackage{siunitx}
\usepackage{multicol}
\usepackage{diagbox}
\usepackage{makecell}
\usepackage{float}

\usepackage{multirow}
\usepackage{booktabs} 
\usepackage{colortbl}
\usepackage{tabularx} 
\usepackage{threeparttable}
\usepackage{ragged2e} 
\usepackage[figuresright]{rotating}
\usepackage{siunitx}

\newcolumntype{C}{>{\centering\arraybackslash}X}
\newcolumntype{L}{>{\raggedright\arraybackslash}X}

\usepackage{cite}
\usepackage{url}
\usepackage{hyperref}
\hypersetup{
colorlinks=true,
linkcolor=blue,
citecolor=blue
}

\usepackage[capitalise,nosort]{cleveref}      

\renewcommand{\eqref}[1]{\mbox{Equation~(\ref{#1})}}

\newcommand{\idel}[1]{}

\usepackage{enumitem}
\newlist{RQlist}{enumerate}{1}
\newlist{RAlist}{enumerate}{1}
\setlist[RQlist]{label=\textbf{R1Q\arabic*:}, leftmargin=*, nosep}
\setlist[RAlist]{label=\textbf{R1A\arabic*:}, leftmargin=*, nosep}
